%% file: iclr2021_workshop.tex
\documentclass{article}
\usepackage{iclr2021_workshop,times}

\input{math_commands.tex}

\usepackage{hyperref}
\usepackage{url}
\usepackage{graphicx}
\usepackage{textcomp}
\usepackage{multirow}
\usepackage{xcolor}
\usepackage{booktabs}
\usepackage{float}
\graphicspath{ {./Figures/} }

\title{GLAMOUR: Graph Learning over Macromolecule Representations}

\author{Somesh Mohapatra, Joyce An \& Rafael Gómez-Bombarelli\thanks{Corresponding author: rafagb@mit.edu.} \\
Department of Materials Science and Engineering\\
Massachusetts Institute of Technology\\
Cambridge, MA 02139, USA \\
\texttt{\{someshm,joycean,rafagb\}@mit.edu}
}

\iclrfinalcopy 
\begin{document}
\maketitle

\begin{abstract}
The near-infinite chemical diversity of natural and artificial macromolecules arises from the vast range of possible component monomers, linkages, and polymers topologies. This enormous macromolecular diversity contributes to the ubiquity and indispensability of macromolecules but hinders the development of general machine learning methods with macromolecules as input. To address this, we developed GLAMOUR, a framework for chemistry-informed graph representation of macromolecules that enables quantifying structural similarity, and interpretable supervised learning for macromolecules. 
\end{abstract}

\section{Introduction}
Machine learning (ML) applications to individual macromolecule classes, such as proteins, have been very successful but typically rely on sequence-based representations that effectively capture structures with linear architectures \citep{Schissel2021, Alley2019}. However, these methods do not extend to macromolecules with complex topologies, such as glycans and biohybrid sequence-defined polymers, which exhibit non-linear structure and higher levels of monomer and linkage diversity. Graphs are a natural and more general macromolecule representation, with nodes representing monomers and edges representing connecting bonds. ML over graph representations has achieved state-of-the-art results across several fields \citep{Hamilton2017} (see Section \ref{si_1} for a survey of the field) and in chemistry and life sciences, graph neural networks (GNNs) have become the modern workhorse for molecular property prediction \citep{Yang2019, Jin2020}. Attribution methods are compatible with GNNs to highlight how input features are relevant to model predictions of target  properties, such as molecular olfactory  descriptors \citep{Sanchez-lengeling2020, Sanchez-lengeling2019}. 

In this work, we developed a graph representation for macromolecules. The representation generalizes ideas of macromolecule similarity from sequence alignment to structural similarity between complex topologies. Using chemical similarity between monomers through cheminformatic fingerprints and exact graph edit distances (GED) or graph kernels to compare topologies, it allows for quantification of the chemical and structural similarity of two arbitrary macromolecule topologies. Further, the representation was coupled to supervised GNN models to learn structure-property relationships in glycans and anti-microbial peptides. The graph representation handles linear, branched and cyclic topologies along with any monomer composition. Attribution methods highlight the regions of the macromolecules and the substructures within the monomers that are most responsible for the predicted properties. 

\section{Results and Discussion}
\begin{figure}[t]
\centering
\includegraphics[width=1\textwidth]{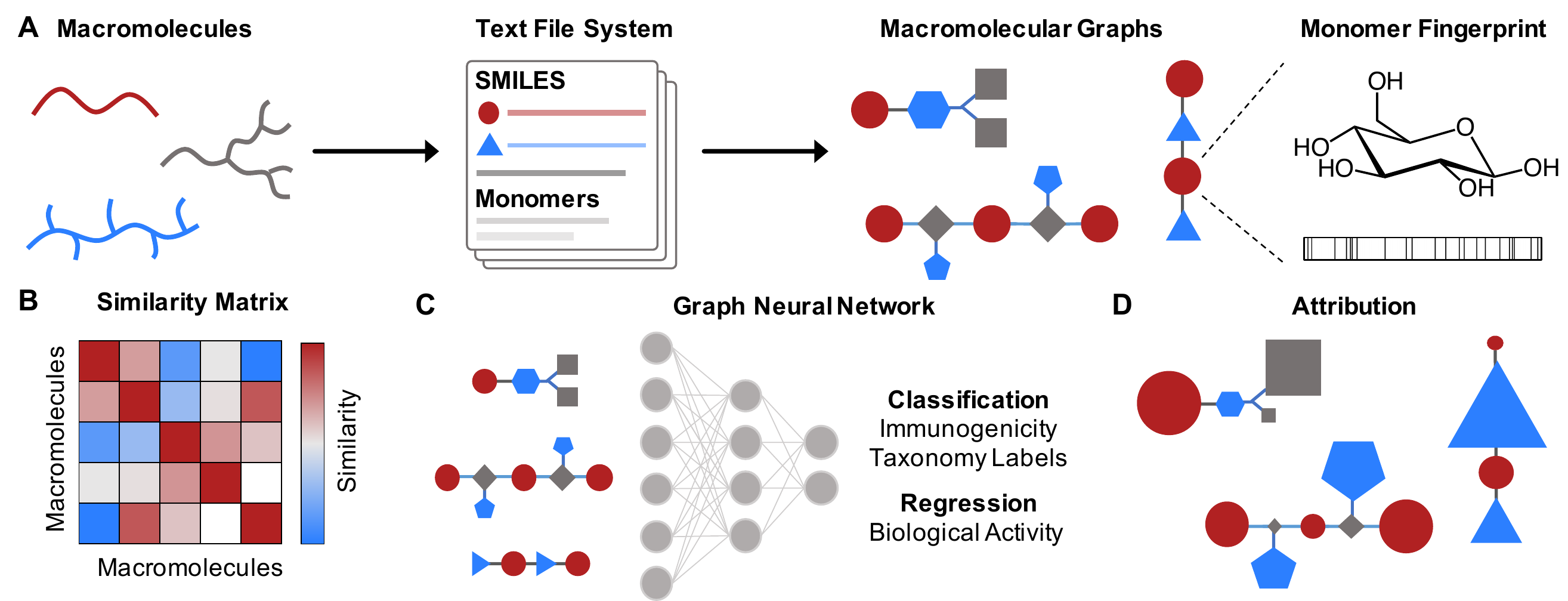}
\caption{\textbf{Macromolecules represented as graphs enable similarity computation and interpretable machine learning.} \textbf{A.} Macromolecular structures are converted into text files. The text files are parsed into network graphs, where monomers are nodes and bonds are edges. Each text file enumerates SMILES for monomers and bonds, node indices corresponding to monomers, and pairs of node indices for bonds. Node and edge attributes are fingerprints of respective molecules. \textbf{B.} Pair-wise similarity matrix obtained for the library of macromolecules, and dimensionality reduction followed by clustering of similarity vectors corresponding to each macromolecule. \textbf{C.} Graph neural networks learn over macromolecule graphs for various tasks. \textbf{D.} Graph attribution provides insight into the model's decision-making process by highlighting the relative importance of different nodes/monomers (denoted as node size) in making a specific prediction.}
\label{fig1}
\end{figure}

We developed a generalized text file system to convert a macromolecule structure into a machine-readable format (Figure \ref{fig1}A, Section \ref{si_3}). The text file has three sections – SMILES, MONOMERS, and BONDS, inspired by the PDB file format \citep{Berman2000}. Under SMILES, monomer and bond names followed by the stereochemical SMILES are noted. MONOMERS enumerates indices of all nodes numbered from 1 to n, where n is the total number of monomers, followed by the monomer names. Similarly, BONDS lists indices of connections between monomer indices, followed by bond names. In this way, we are able to incorporate complexity from the level of individual atoms to the full macromolecular structure. 

In our experiments, the macromolecules were then processed as NetworkX graphs \citep{Hagberg2008}, with monomers as nodes and bonds as edges (Section \ref{si_4}). In line with earlier work where fingerprint-based monomer representations worked effectively for macromolecule property prediction, the monomer and bond molecules were featurized using standard Extended Connectivity FingerPrints \citep{Schissel2021}. The fingerprints capture the atomic connectivity of the monomer/bond molecule as a series of bits using circular atom neighborhoods, for each constituent node or edge of the macromolecule graph. This representation enables the encoding of the raw chain connectivity of macromolecules with explicit featurization of the stereochemistry and topology. Moreover, it provides a single framework to represent both natural and synthetic, linear and non-linear macromolecules.

\begin{figure}[t]
\centering
\includegraphics[width=1\textwidth]{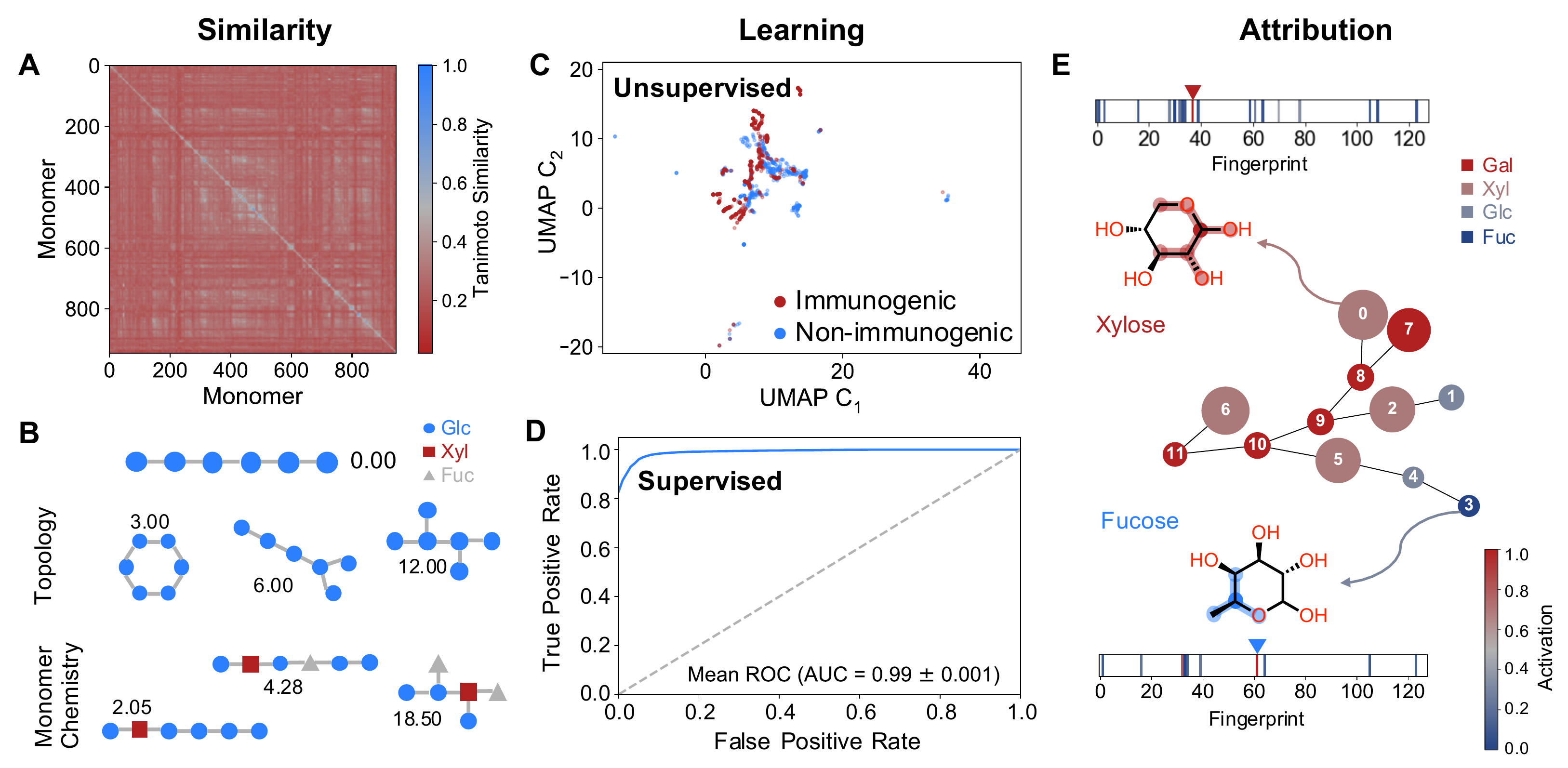}
\caption{\textbf{Analysis using GLAMOUR provides deeper insights into immunogenicity of glycans.} \textbf{A.} Pairwise chemical similarity between glycan monomers using Tanimoto distance on ECFP fingerprints is typically low. \textbf{B.} Similarities between a linear glycan consisting of six glucose monomers with itself and six additional glycans with modified topology and monomer chemistry were computed using exact graph edit distances, combined with the Tanimoto substitution matrices (A). \textbf{C.} 2-component UMAP over the similarity matrix, colored by immunogenicity, for 1313 glycans with immunogenicity labels. \textbf{D.} Mean ROC-AUC curve for classification of immunogenic and non-immunogenic glycans, averaged across all 25 Attentive FP models (5 random seeds, with top 5 sets of hyperparameters), with the standard deviation shaded in light blue too insignificant to be visible in the graph. \textbf{E.} Attribution analysis using Attentive FP model architecture and integrated gradients attribution method for a representative glycan shows that xylose (Xyl) contributes the most to immunogenicity. In xylose, the substructure centered on the anomeric carbon, C1, is shown to be the key contributor. A similar substructure analysis has been shown for fucose. In the glycan graph, the size of the node corresponds to the importance, and the color corresponds to the monomer. In the fingerprint, only the indices with positive contribution to immunogenicity have been visualized, with the color corresponding to the normalized importance}
\label{fig2}
\end{figure}

To compute the similarity between 2 or more macromolecule graphs, we used GED and graph kernel (Figure \ref{fig1}B, Section \ref{si_5}). GED \citep{abuaisheh:hal-01168816} computes the similarity between two graphs by assigning node and edge substitution scores, similar to local sequence alignment methods, such as BLAST \citep{Altschul1990}. Instead of evolutionary statistics-based substitution matrices like BLOSUM62, we used Tanimoto chemical similarity matrices that compute the similarity between molecular fingerprints. Tanimoto similarity also extends to unnatural monomers. Since computing exact graph edit distances is an NP-hard problem, we used propagation attribute graph kernels to obtain approximate similarity matrices for large datasets \citep{Neumann2016, Siglidis2020}. This graph kernel captures local monomer node information and propagates this information along the bond edges, making it an ideal choice for macromolecule graphs. We have demonstrated the similarity computation for a linear glycan with six additional glycans of different topology and/or monomer chemistry, as well as with itself, using Tanimoto chemical similarity matrix (Figure \ref{fig2}A, B).

Because quantifying graph similarity enables unsupervised learning over similarities rather than raw features, we used dimensionality reduction methods, such as uniform manifold approximation and projection (UMAP) \citep{McInnes2018} and t-stochastic neighbor embeddings (t-SNE) \citep{VanDerMaaten2008VisualizingT-SNE}, over the similarity matrices for unsupervised learning (Section \ref{si_6}). For glycans with immunogenicity labels, we observed that the non-immunogenic and immunogenic glycans are in nearly distinct regions (Figure \ref{fig2}C).

For supervised learning, we trained five different GNN model architectures to classify glycans by immunogenicity and taxonomy levels, and predict the anti-microbial activity of peptides (Section \ref{si_7}). Each model architecture was trained over fingerprint and one-hot node and edge attributes on 60\%, validated on 20\%, and tested on held-out 20\% data set, all determined via random splits. Hyperparameter optimization was carried out for 1000 iterations using SigOpt \citep{sigopt-web-page}. To report final metrics on validation and test datasets, for each model architecture, we averaged values obtained from 25 individual models – top 5 hyperparameter sets, each trained with 5 random seeds. For glycans, our models outperformed metrics for classification for four out of eight tasks and achieved comparable results for the rest four, against results reported in the literature (Figure \ref{fig2}D, Table \ref{sitab5}) \citep{Burkholz2021}.

Attribution analysis on the pre-trained models finds features – both nodes and substructures – that are key to the model’s decision-making (Section \ref{si_8}). To avoid spurious hypotheses \citep{McCloskey2019} that may occur in NN attribution on molecular models, we chose the combination of GNN architecture and attribution method, in this case Attentive FP - integrated gradients, that produced the most consistent results across all immunogenic glycans over 25 replicates (Figure \ref{sifig45}). N-glycolylneuraminic acid, xylose, and fucose were found to be the key monomers responsible for immunogenicity, consistent with experimental findings due to the low prevalence of these monomers in human glycans \citep{Planinc2016} (Table \ref{sitab7}). For a representative glycan, we observed that xylose, followed by galectin, were the monomers that contributed most significantly to immunogenicity (Figure 2E). In addition to the importance of individual nodes, attribution identifies the critical substructures in the monomers that contributed most to immunogenicity, such as the substructure centered on the anomeric carbon of xylose. These critical features help understand the fundamental structure-function relationships that underpin immunogenicity, postulate very explicit hypotheses that can be validated in the lab, and may help in further design of immunogenic or non-immunogenic scaffolds.

\section{Conclusion}
GLAMOUR provides a generalized method for representing macromolecules as hierarchical graphs with molecular fingerprints to capture chemical information which can be used to compute structural similarity between macromolecules with different composition and topology, and perform unsupervised and supervised learning. The unsupervised learning enables visualization of the complex landscape of different classes of macromolecules and understanding of the subtle differences between similar macromolecules. The attribution analysis helps in cracking open the black-box of supervised GNN models, which in turn can help elucidate fundamental design principles of otherwise opaque structure-property relationships and assist with hypothesis generation for future experimental studies. The codebase and data sets have been open-sourced \citep{somesh_mohapatra_2021_5237237}. We therefore expect that this toolkit will be used by both experimentalists and computational practitioners in chemistry, biology and materials science for a variety of macromolecule property prediction tasks, ultimately driving the design of improved macromolecules. 

\section*{Data availability}
Data used for training of the model is available at \href{https://github.com/learningmatter-mit/GLAMOUR}{https://github.com/learningmatter-mit/GLAMOUR}, and archived in the Zenodo repository \citep{somesh_mohapatra_2021_5237237}.

\section*{Code availability}
All the code used for model training and analysis is available at \href{https://github.com/learningmatter-mit/GLAMOUR}{https://github.com/learningmatter-mit/GLAMOUR}, and archived in Zenodo repository \citep{somesh_mohapatra_2021_5237237}.

\section*{Acknowledgements}
SM acknowledges funding from Novo Nordisk and MIT’s Jameel Clinic, JA acknowledges funding from MIT’s UROP office. The authors thank Dr Daria Kim and Sarah Antilla from the Department of Chemistry, MIT for valuable comments.

\vfill
\pagebreak

\bibliography{iclr2021_workshop}
\bibliographystyle{iclr2021_workshop}
\vfill
\pagebreak

\appendix
\renewcommand\thefigure{A.\arabic{figure}}
\setcounter{figure}{0}    

\renewcommand\thetable{A.\arabic{table}}
\setcounter{table}{0}

\LARGE A\Large PPENDIX
\normalsize
\section{Related work}
\label{si_1}
\emph{Representation.} Macromolecules can be represented as strings, using hierarchical editing language for macromolecules (HELM) \citep{Zhang2012}, International Union of Pure and Applied Chemistry (IUPAC) international chemical identifier (InChI) \citep{Heller2015}, CurlySMILES \citep{Drefahl2011} (where SMILES is Simplified Molecular-Input Line-Entry System) and BigSMILES \citep{Lin2019a}. Line notations do not always support all topologies, require a fair amount of customization, and have non-canonical variants. One exception to this is linear biological macromolecules, such as proteins and DNA/RNA, which are represented as sequences of one/three-letter monomer codes. In a recent attempt, glycans (non-linear biological macromolecules) were represented as sequences, where groups of monosaccharides were clubbed into 'glycowords' and placed in hierarchical brackets \citep{Bojar2021}. Representing glycans presents a significant advance in notation but falls short in terms of comparison (limited to linear glycans) and machine learning tasks.

Hierarchical fingerprinting is another approach, which follows a hierarchy of atomic (categorical encoding of presence/absence of contiguous atom-sets), physicochemical (e.g., fraction of rings, molecular surface area), and morphological descriptors (e.g., length of side chain, length of main chain) \citep{Kim2018}. This approach is limited in its coverage of chemical space, ability to differentiate stereochemistry, and capturing long range through-space interactions.

\emph{Similarity computation.} In recent times, there have been significant advances in computation of similarity using graph edit distances \citep{Blumenthal2020} (GED) and graph kernels \citep{Borgwardt2020, kriege2020survey}. Development of software packages, such as graphkernels \citep{Sugiyama2018} and GraKeL \citep{Siglidis2020}, has provided fast implementations of graph kernels.

For linear biological macromolecules, such as proteins, DNA/RNA and linear glycans, there are several works for computation of sequence similarities \citep{Bojar2021, Altschul1990, Altschul1997, Boratyn2019}. Usually, sequence alignment is done using Smith-Waterman \citep{Smith1981} or Needleman-Wunsch \citep{Needleman1970} algorithms, and scored with substitution matrices, such as BLOSUM62 \citep{Eddy2004} (for proteins) and GLYSUM \citep{Bojar2021} (for glycans). These substitution matrices are based on evolutionary statistics thereby biasing the scoring towards the statistical frequency of a particular monomer's occurrence in the course of evolution, rather than chemical similarity. Apart from sequence alignment in linear macromolecules, edit distances \citep{Riesen2009, Zhang2010}, linear kernels \citep{Leslie2001, Jaakkola2000} and deep learning methods \citep{Bileschi2019, Seo2018} have been proposed to compute similarity. In the case of non-linear macromolecules, alignment of glycans has been explored using q-grams \citep{Li2010}, tree matching methods \citep{Aoki2003, Hosoda2017, Coff2020}, and using tree kernels \citep{Yamanishi2007}. Unfortunately, the aforementioned methods are limited to biological macromolecules, and do not extend to the general macromolecular chemical space. Moreover, existing tools do not allow incorporation of unnatural monomers and non-linear topologies (except glycans).

\emph{Machine learning.} The field of graph neural networks (GNN) has seen substantial developments in both model architecture and attribution. The different model architectures have demonstrated state-of-the-art results across various domains - graph convolutional network (GCN), graph attention network (GAT), message passing neural network (MPNN), SchNet, Weave (a variant of GCN), and AttentiveFP (a variant of GAT) \citep{Zhou2019, Schutt2017schnet, Kearnes2016, Xiong2020}. Graph attribution has been recently studied quantitatively across four metrics – accuracy, stability, faithfulness and consistency - for a wide variety of tasks and model architectures \citep{Sanchez-lengeling2020}.

For macromolecular property prediction, Polymer Genome and similar works using hierarchical fingerprints predict glass transition temperature, dielectric point and other macromolecular properties \citep{Kim2018, Chen2021}. There have been several attempts to extrapolate macromolecular property by training over monomer input features \citep{StJohn2019, Qiao2020a, Lee2021}. GCN over macromolecule graphs with one-hot node and edge attributes has been shown to outperform fingerprint-based models \citep{Zeng}. For fingerprint-based models, the lack of a good macromolecule representation limits the model performance, while the GCN model is limited by the lack of chemical information in the graph representation.

\section{Dataset processing}
\label{si_2}
\subsection{Glycans}
\label{si_2_1}
\subsubsection{Dataset download}
\label{si_2_1_1}
A dataset of 19299 glycans was accessed and downloaded from GlycoBase (accessed on November 2, 2020) \citep{Bojar2021}. The file contained GlycoBase ID, sequence, link (N, O, free, or none), species, and immunogenicity information for each glycan. 

For each glycan sequence string the brackets denote branches, with the point of attachment/bonding of the branch as the monomer immediately after the brackets. The first element within the bracket is the monomer most distant from the point of attachment, and the last element within the bracket is the abbreviation of the bond that connects the branch to the original main chain. Nested brackets indicate additional sub-branches off of branches, and multiple sets of brackets next to each other indicate several branches off of the same monomer. 

\subsubsection{Dataset pre-processing}
\label{si_2_1_2}
7 modifications and 152 deletions of glycan sequences were made before the process for situations such as an unequal number of opening and closing brackets and dangling branches without specified connectivity. Additional glycan sequences were removed due to missing SMILES sequences for a number of monomers. The original GlycoBase.csv file was curated to reflect the modification and deletion changes, resulting in a total of 19147 glycans. 

\begin{figure}[h!]
\centering
\includegraphics[width=1\textwidth]{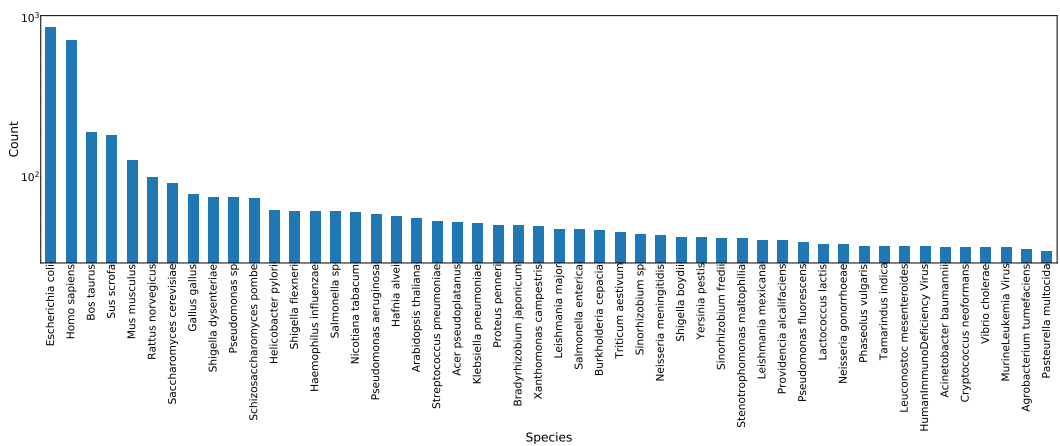}
\caption{Top 50 most common glycan species of origin, sorted in descending order of count with y-axis on a logarithmic scale.}
\label{sifig1}
\end{figure}

\begin{figure}[h!]
\centering
\includegraphics[width=1\textwidth]{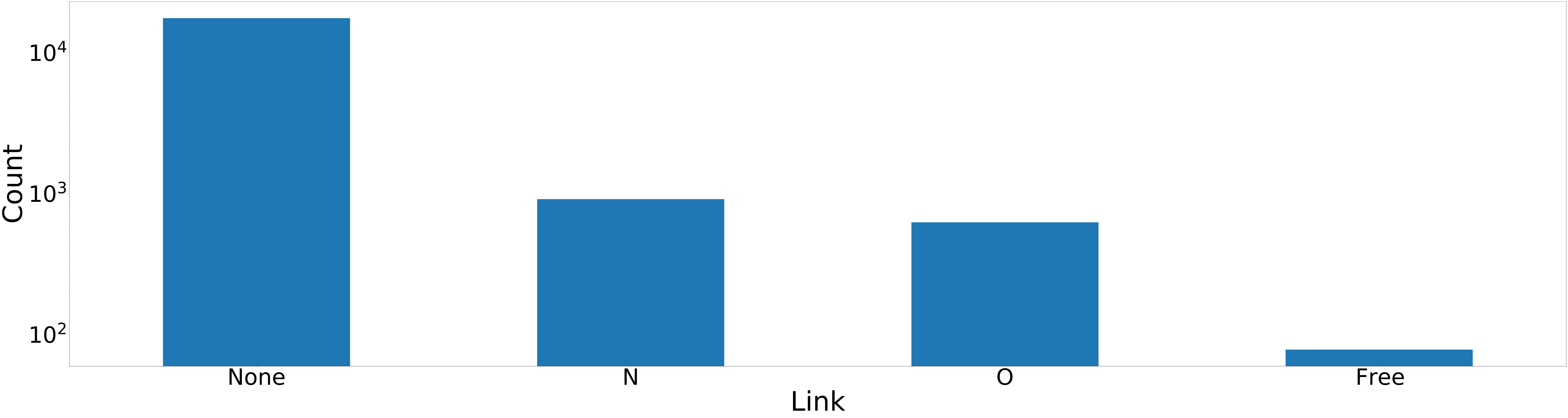}
\caption{Four types of glycan links sorted in descending order of count with y-axis on a logarithmic scale.}
\label{sifig2}
\end{figure}

\begin{figure}[h!]
\centering
\includegraphics[width=1\textwidth]{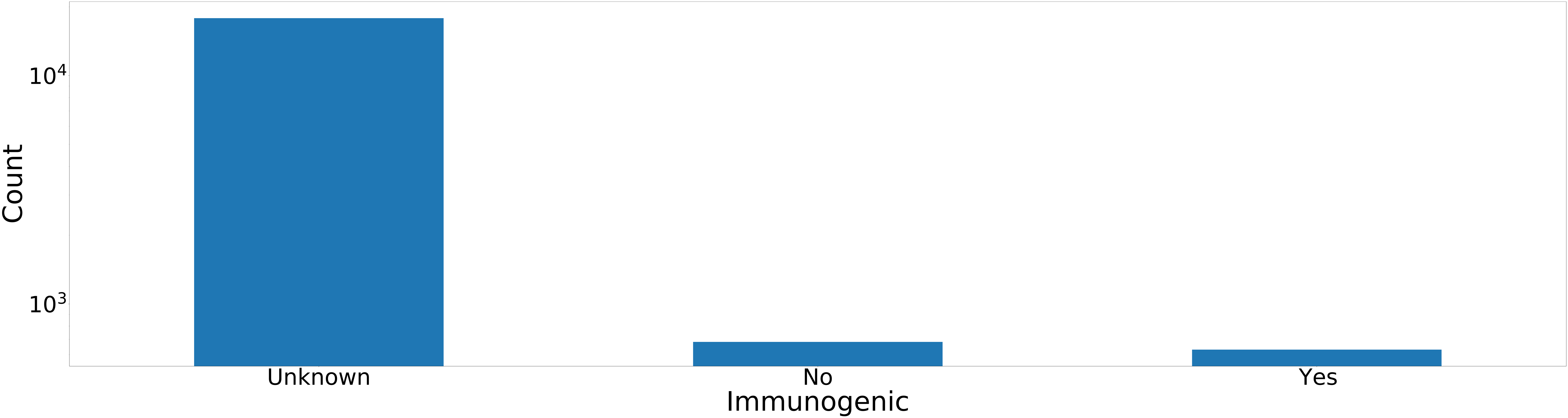}
\caption{Three different types of glycan immunogenicity labels sorted in descending order of count, with y-axis on a logarithmic scale.}
\label{sifig3}
\end{figure}

Using the curated database, we visualized the distribution of species of origin, link, and immunogenicity of the glycans (Figures \ref{sifig1}, \ref{sifig2}, \ref{sifig3}).

\subsubsection{SMILES compilation}
\label{si_2_1_3}
The chemical composition and formula of the 959 monomers (also known as glycoletters in GlycoBase) and 53 bonds were expressed as isomeric simplified molecular-input line-entry system (SMILES) sequences. 

In the dataset retrieved from GlycoBase, all position-specific information about monomer modifications was removed. The supplementary information of SweetTalk contains all the raw glycan sequences before position-specific modification information was removed \citep{Bojar2021}. While insufficient information is provided to directly match the raw sequences with the corresponding edited sequences, the raw sequences can be used to look for trends and most common modification positions for each monomer. A dictionary was created to describe the number of times each monomer appears in all the raw sequences. A couple of terms should be defined for consistency:
\begin{itemize}
  \item “Monosaccharides” refer to the individual glycoletters without any modifications, such as D-glucose (D-Glu) and galactose (D-Gal). The term “monosaccharide” is simplistic because it also covers alcohols, acids, and other classes of molecules, but the term will be employed for the sake of consistency and clarity.
  \item “Modifications” refer to substitutions or additions to the original monosaccharide sequence such as a nitrogen-linked acetyl group (NAc) and oxygen-linked methyl group (OMe). In this dataset, monosaccharides can have between 0-4 modifications.
  \item “Monomers” refer to the combination of the monosaccharide and modification(s), also known as glycoletters in GlycoBase.
\end{itemize}

Using the raw monomers dictionary, the positions of the modifications for each monomer in the corresponding SMILES were set using a list of consistent rules outlined below. The rules are listed in order of priority.

\begin{enumerate}
  \item For each monomer/glycoletter, if at least one monomer with the same monosaccharide and set of modifications exists in the raw monomers dictionary, assume the set of positions with the highest frequency. If no monomer exists in the raw monomers dictionary with the same monosaccharide and set of modifications, proceed to the following steps. 
\item If the monomer has a hydroxyl group at the 2 position, is not a furanose or ketose, and contains a nitrogen-linked modification (N, NBut, NMe, etc.), assign the first occurrence of a nitrogen-linked modification to the 2 position. 
\item If the monomer is a hexose or deoxy-hexose and contains a pyruvate acetal (OPyr), assume that the acetal connects the 4 and 6 positions.
\item If the monomer contains an O-linked phosphate (OP) or sulfate (OS) modification, search the raw monomers dictionary for instances of the monosaccharide with only the OP/OS modification and assume the most frequent position.  
\item If the monomer contains a variant of the O-linked phosphate modification (OPEtn, OPPEtn, etc.), assign the OP-variant modification the same position as the most common position for the OP modification on the monosaccharide. 
\item If the monomer contains two modifications, search the raw glycans dictionary for instances of the monosaccharide with only each modification separately but not at the same time. Assign each modification the position of highest frequency. If either monosaccharide and modification combination does not exist in the raw monomers dictionary and the monomer is a hexose, assign positions in the following order: 2, 4, 3, 6. If the position is already occupied from previous steps or does not exist in a deoxy hexose, skip to the position with next highest priority. 
\item If the monomer is a hexose or deoxy-hexose and contains more than two modifications, assign positions in the following order: 2, 3, 4, 6. If the position is already occupied from previous steps or does not exist in a deoxy hexose, skip to the position with next highest priority. 
\item Assume all amino acids are connected to the monosaccharide via the oxygen on the carboxyl. If the connectivity is not specified for a group following the amino acid (CysAc, AlaFo), assume that the group following the amino acid is connected to the amino acid via the amine group. 
\item For neuraminic acid (Neu), ketodeoxynononic acid (Kdn), pseudaminic acid (Pse), legionaminic acid (Leg), and other similar ketose-based acids, assign modification positions in the following order: 1, 4. 
\item For fructofuranose and similar furanoses for which the 1-position is not part of the ring, assign modifications in the following order: 1, 3, 4. 
\item For alcohols, follow the same rules that apply to the oxidized form of the alcohol. For example, for glucitol follow the same rules that apply to glucose. 
\item Assume all rare monosaccharides (denoted as Sug) are hexoses with no specified stereochemistry. 
\end{enumerate}

Because the 959 monomers contained some repeat SMILES sequences with different monomer names, 13 redundant names were deleted so that each distinct monomer SMILES only appears once in the SMILES compilation. 

The SMILES for the 53 bond types differ in stereochemistry alone (alpha, beta, or unspecified) but all have the same chemical composition of a glycosidic bond. Each bond is expressed as a variation of the SMILES sequence CC(OC)CC, which consists of the glycosidic bond C-O-C with one of the C atoms also connected to both a methyl and ethyl group. The chiral C with the four different attached groups is used to specify the alpha or beta stereochemistry displayed in the bond name. The stereochemistry at the tetrahedral C is consistently S for all alpha bonds, R for all beta bonds, and not specified for all unspecified bonds. The 53 bond names were condensed into 3 distinct bond types differing in stereochemistry alone. 

\subsection{Anti-microbial peptides}
\label{si_2_2}
\subsubsection{Dataset download}
\label{si_2_2_1}
A dataset of 15864 antimicrobial peptides (AMPs), including 15450 monomers, 200 multimers, and 214 multi-peptides, from the Database of Antimicrobial Activity and Structure of Peptides (DBAASP) was accessed and downloaded on October 6, 2020 \citep{Pirtskhalava2016}. 

Each peptide was represented in the dataset as an individual .json file containing information about the peptide ID, name, sequence(s), unusual amino acids, connectivity, terminal modifications, complexity, synthesis type, target groups and objects, and target species. The term “target species” is used loosely and actually encompasses both species and sub-classification information such as subspecies, strain for bacteria, and forma speciales for fungi. For each target species for any given peptide, the dataset provides the antimicrobial peptide concentration in units of mostly either $\mu$M or $\mu$g/mL as a function of four unique variables: activity measure, salt type, medium, and CFU. 

The dataset includes three different types of peptide complexities: monomers, multimers, and multi-peptides. In the DBAASP dataset, monomers consist of a single sequence, multimers between 2 and 4 separate sequences connected via interchain bonds, and multi-peptides between 2 and 4 separate sequences connected not via covalent bonds but instead weaker intermolecular forces.

\subsubsection{Dataset pre-processing}
\label{si_2_2_2}
The information from each .json file was inputted into a combined table and the antimicrobial peptide concentrations for each target species was converted into a float value by removing symbols like ‘$>$’ and ‘$<$’, taking the average whenever the dataset provides a range, and disregarding uncertainty values. 

9 peptide monomer types included in a total of 86 peptides were removed due to ambiguity of the molecular structure, bringing the total number of peptides to 15,778. This condensed dataset was further processed to visualize the distribution of target species data points. For each “target species,” the species name was separated from any sub-classifications (subspecies, strain, forma speciales, serovar, pathovar, biovar, etc.). 

\begin{figure}[h!]
\centering
\includegraphics[width=1\textwidth]{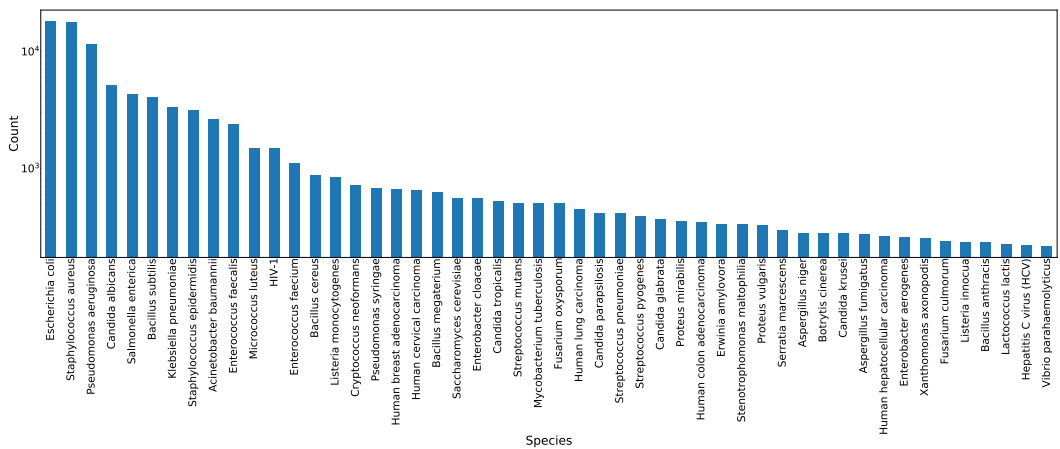}
\caption{Top 50 species with greatest number of data points sorted in descending order of count, with y-axis on a logarithmic scale.}
\label{sifig4}
\end{figure}

\begin{figure}[h!]
\centering
\includegraphics[width=1\textwidth]{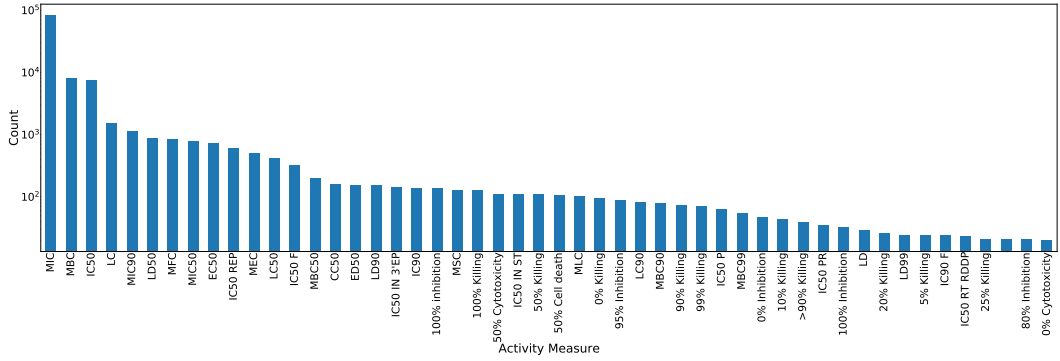}
\caption{Top 50 activity measures with greatest number of data points sorted in descending order of count, with y-axis on a logarithmic scale.}
\label{sifig5}
\end{figure}

Using the processed database, we visualized the distribution of species, activity measure, salt types, mediums, units, and CFU in the target species data points (Figures \ref{sifig4}-\ref{sifig9}). 

\begin{figure}[h!]
\centering
\includegraphics[width=1\textwidth]{Figures/SIFig6.pdf}
\caption{Top 50 salt types with greatest number of data points sorted in descending order of count, with y-axis on a logarithmic scale. The blank label for the first column indicates the absence of any added salt.}
\label{sifig6}
\end{figure}

\begin{figure}[h!]
\centering
\includegraphics[width=1\textwidth]{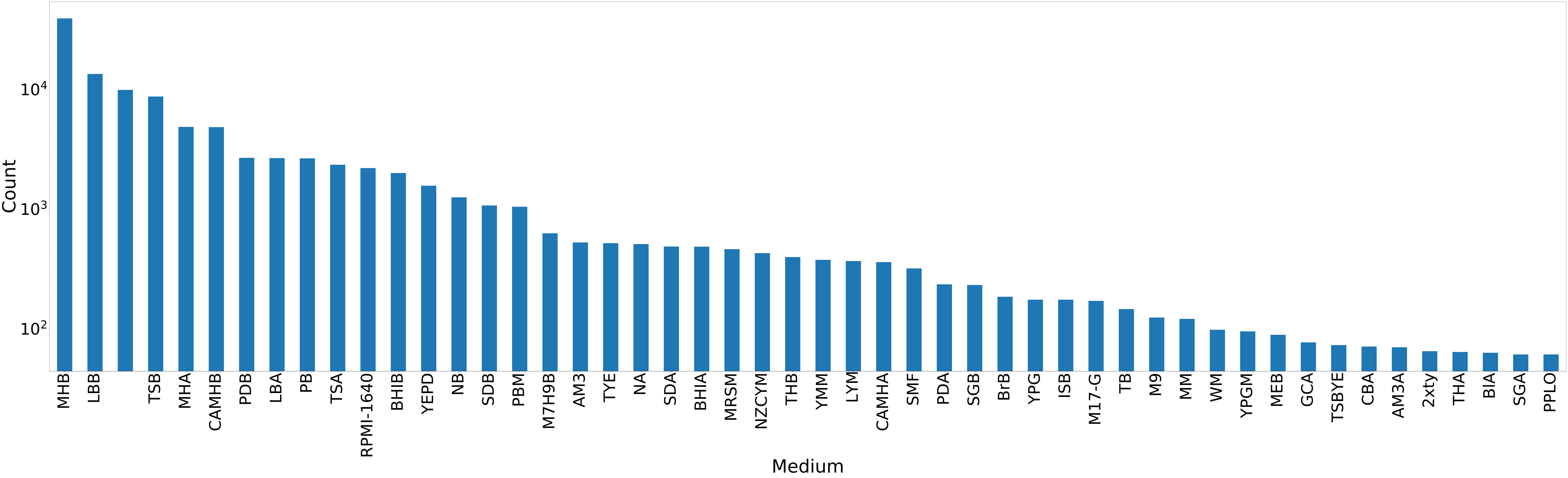}
\caption{Top 50 mediums with greatest number of data points sorted in descending order of count, with y-axis on a logarithmic scale.}
\label{sifig7}
\end{figure}

\begin{figure}[h!]
\centering
\includegraphics[width=1\textwidth]{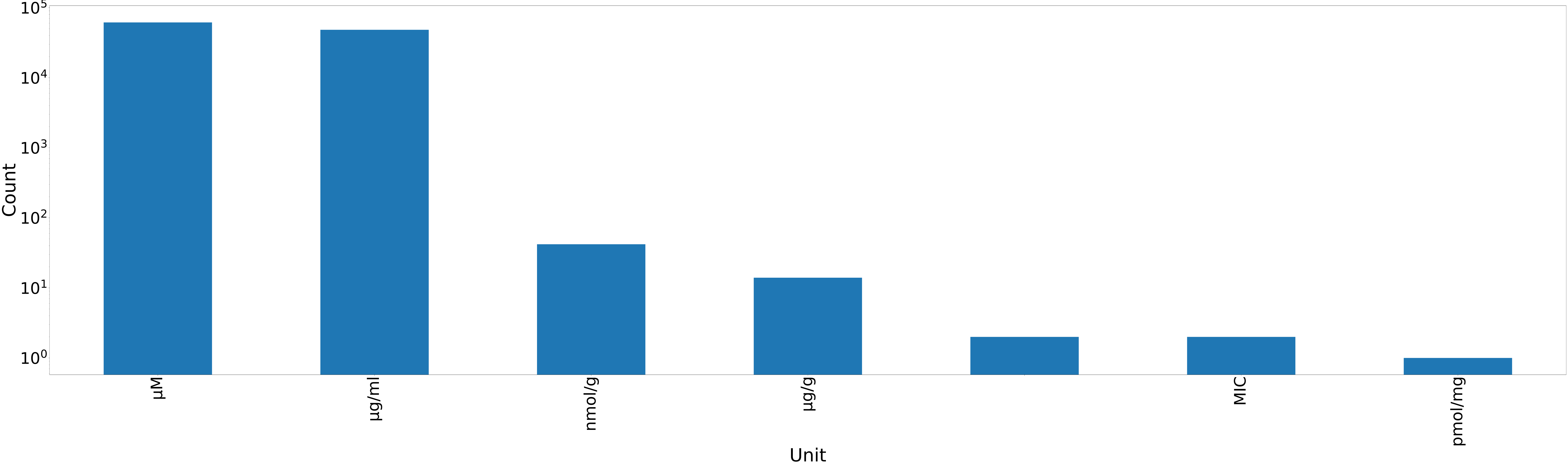}
\caption{All peptide concentration units sorted in descending order of count, with y-axis on a logarithmic scale.}
\label{sifig8}
\end{figure}

\begin{figure}[h!]
\centering
\includegraphics[width=1\textwidth]{Figures/SIFig9.pdf}
\caption{Top 50 CFUs with greatest number of data points sorted in descending order of count, with y-axis on a logarithmic scale.}
\label{sifig9}
\end{figure}

As evident from the bar plots, the vast majority of data points belong to either \emph{Escherichia coli} or \emph{Staphylococcus aereus}. To understand whether the overall dataset could be further filtered to limit the variability in conditions for more meaningful comparison of peptide concentrations, we visualized the distribution of activity measure and salt type for both \emph{E. coli} (Figures \ref{sifig10}, \ref{sifig11}) and \emph{S. aereus} (Figures \ref{sifig12}, \ref{sifig13}). 

\begin{figure}[h!]
\centering
\includegraphics[width=1\textwidth]{Figures/SIFig10.pdf}
\caption{Top 50 activity measures for \emph{E. coli} data points with greatest number of occurrences sorted in descending order of count, with y-axis on a logarithmic scale.}
\label{sifig10}
\end{figure}

\begin{figure}[h!]
\centering
\includegraphics[width=1\textwidth]{Figures/SIFig11.pdf}
\caption{Top 50 salt types for \emph{E. coli} data points with greatest number of occurrences sorted in descending order of count, with y-axis on a logarithmic scale. The blank label for the first column indicates the absence of any added salt.}
\label{sifig11}
\end{figure}

\begin{figure}[h!]
\centering
\includegraphics[width=1\textwidth]{Figures/SIFig12.pdf}
\caption{Top 50 activity measures for \emph{S. aereus} data points with greatest number of occurrences sorted in descending order of count, with y-axis on a logarithmic scale.}
\label{sifig12}
\end{figure}

\begin{figure}[h!]
\centering
\includegraphics[width=1\textwidth]{Figures/SIFig13.pdf}
\caption{Top 50 salt types for \emph{S. aereus} data points with greatest number of occurrences sorted in descending order of count, with y-axis on a logarithmic scale. The blank label for the first column indicates the absence of any added salt.}
\label{sifig13}
\end{figure}

For both \emph{E. coli} and \emph{S. aereus}, the majority of data points have an activity measure of MIC and no added salt. Due to the clear majority for both activity measure and salt type as well as the difficulty of interconversion between the units of $\mu$M and $\mu$g/mL, the overall dataset was filtered into two separate, smaller datasets: one containing only data points falling into all of the categories of \emph{E. coli}, MIC, no salt, and $\mu$M data points and the other containing only data points falling into all of the categories of \emph{S. aereus}, MIC, no salt, and $\mu$M. Each of the two sub-datasets was further filtered to remove any peptides for which the standard deviation of the peptide concentrations meeting the criteria for the sub-dataset was not zero, resulting in a total of 4,445 peptides for the \emph{E. coli} dataset and 3,686 peptides for the \emph{S. aereus} dataset. 

\subsubsection{SMILES compilation}
\label{si_2_2_3}
The chemical composition and formula of the 332 amino acids, 93 N-termini modifications, 37 C-termini modifications, and 22 bonds were expressed as isomeric simplified molecular-input line-entry system (SMILES) sequences. The majority of isomeric SMILES sequences were found on PubChem from the National Library of Medicine \citep{Kim2019pubchem}. The remaining SMILES were collected through review of literature cited in the DBAASP peptide cards of peptides containing the corresponding monomers or bonds. 

For the monomers, no additional sequences beyond those of the 20 standard L-amino acids and unusual amino acids, C-termini modifications, and N-termini modifications found in the Help section of the DBAASP website were included in the SMILES compilation. 8 monomers had repeat names between the three different types of monomers, and the names were changed so that each monomer corresponds with a unique name. For the bonds, the sidechain-mainchain bonds (SMB) were split into 7 subtypes determined through review of the literature. In addition, 6 additional bond types were added for cases in which the connection between the N or C-termini modification and terminal amino acid is not an amide bond. The 7 SMB subtypes, 6 additional terminal bond types, and 9 intrachain bond types in the Help section of the DBAASP website resulted in a total of 22 bonds. 

\section{Text file system}
\label{si_3}
\subsection{Format}
\label{si_3_1}
\begin{figure}[h]
\centering
\includegraphics[width=1\textwidth]{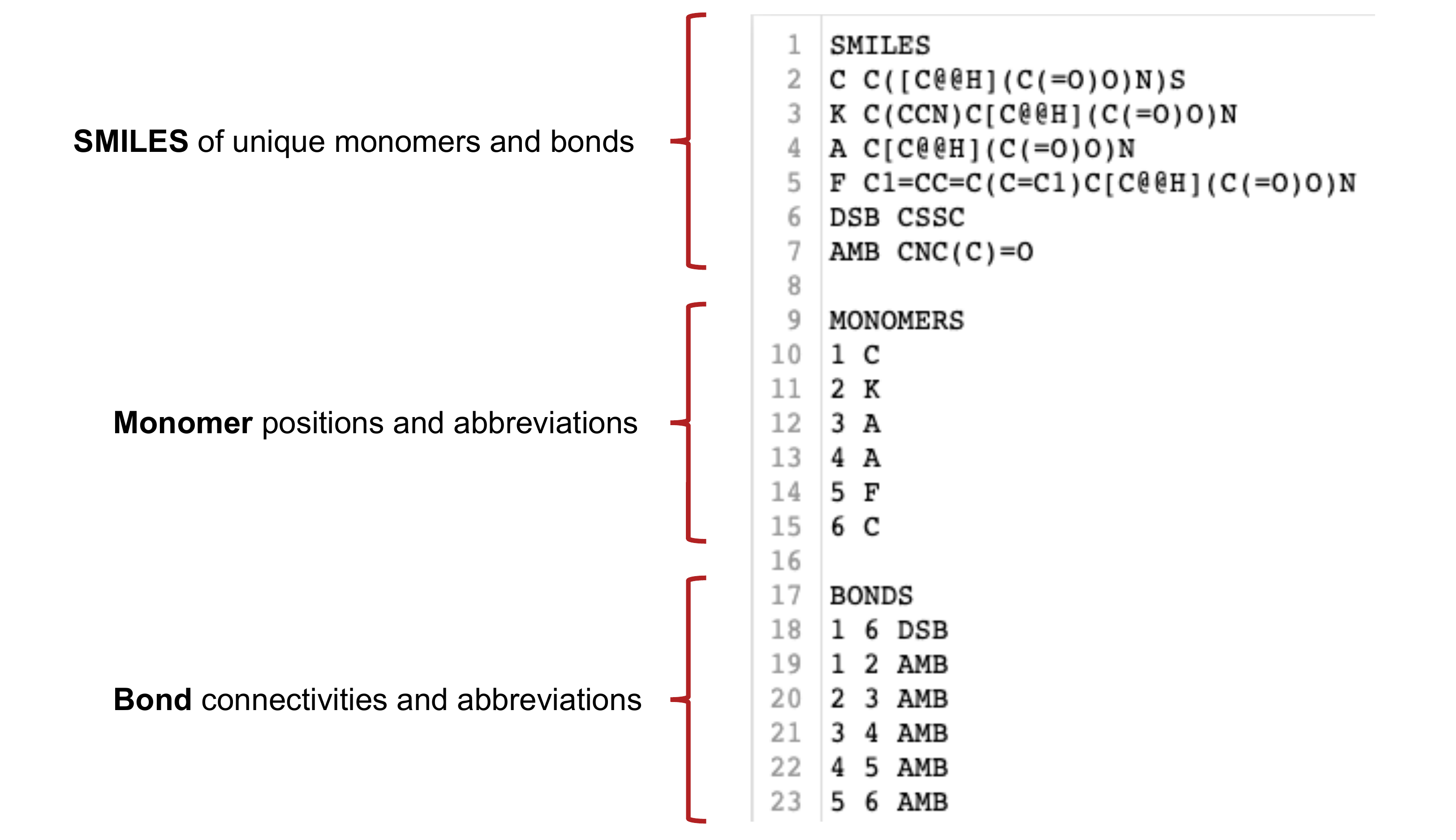}
\caption{Standard text file format with three sections: SMILES of unique monomers and bonds, monomer positions and abbreviations, and bond connectivities and abbreviations.}
\label{sifig14}
\end{figure}

The text files to convert each macromolecule structure into machine-readable format consist of three sections: the SMILES sequences for each unique monomer or bond, the positions of each monomer, and the two monomer positions connected by each bond. Each section starts with a header to indicate the start of a new part. The first section contains the abbreviation of each unique monomer or bond in the glycan followed by the corresponding SMILES sequence, with each entry on a separate line. The monomers section consists of the monomer position followed by the monomer abbreviation. The bonds section contains the two connectivity positions in the glycan for the bond separated by a space, followed by the bond abbreviation (Figure \ref{sifig14}). 

\subsection{Text File Parser}
\label{si_3_2}
\begin{figure}[h]
\centering
\includegraphics[width=1\textwidth]{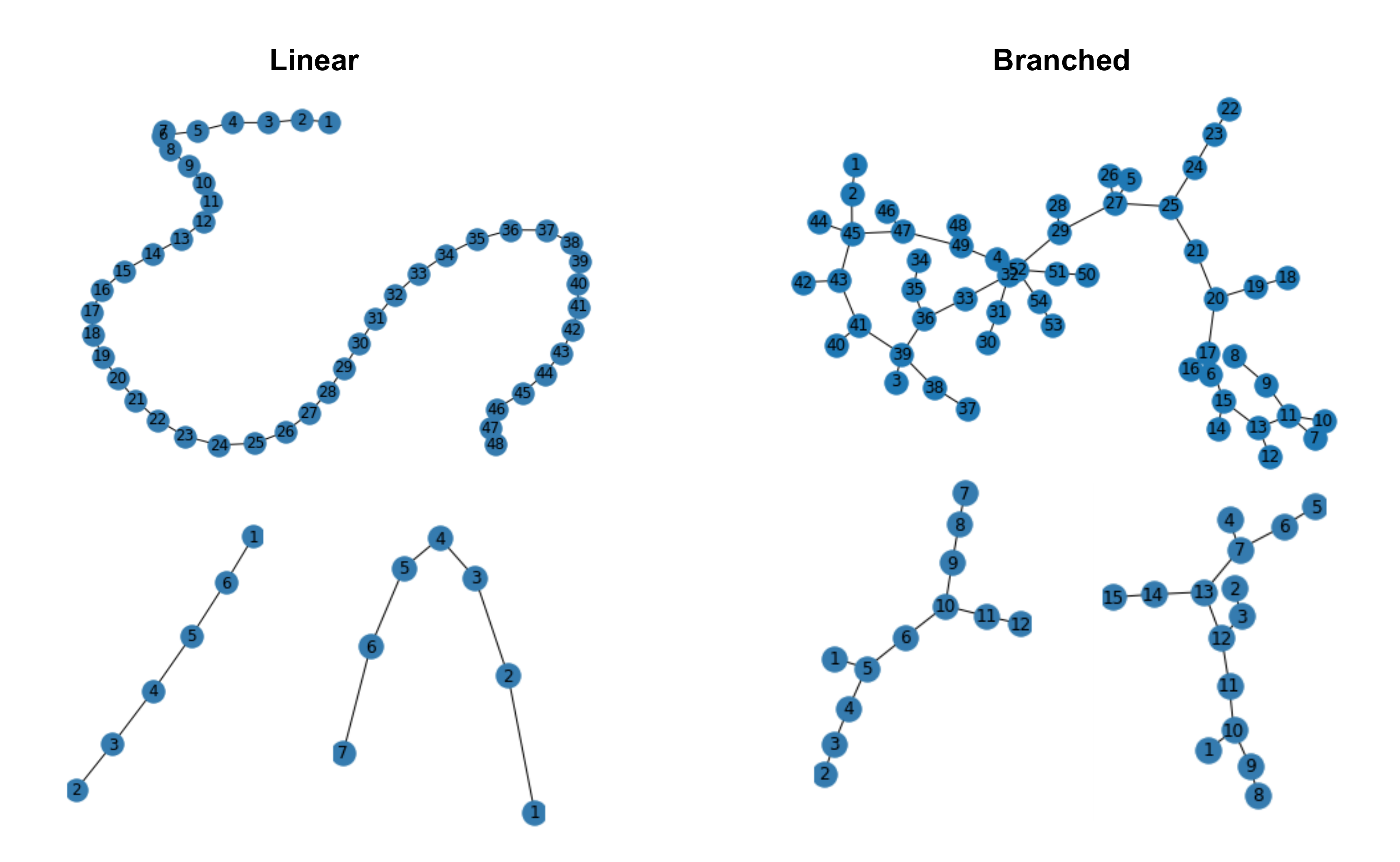}
\caption{NetworkX graph representations of glycans that consist of both linear and highly branched architectures of varying molecular weights and complexities.}
\label{sifig15}
\end{figure}

\begin{figure}[h]
\centering
\includegraphics[width=1\textwidth]{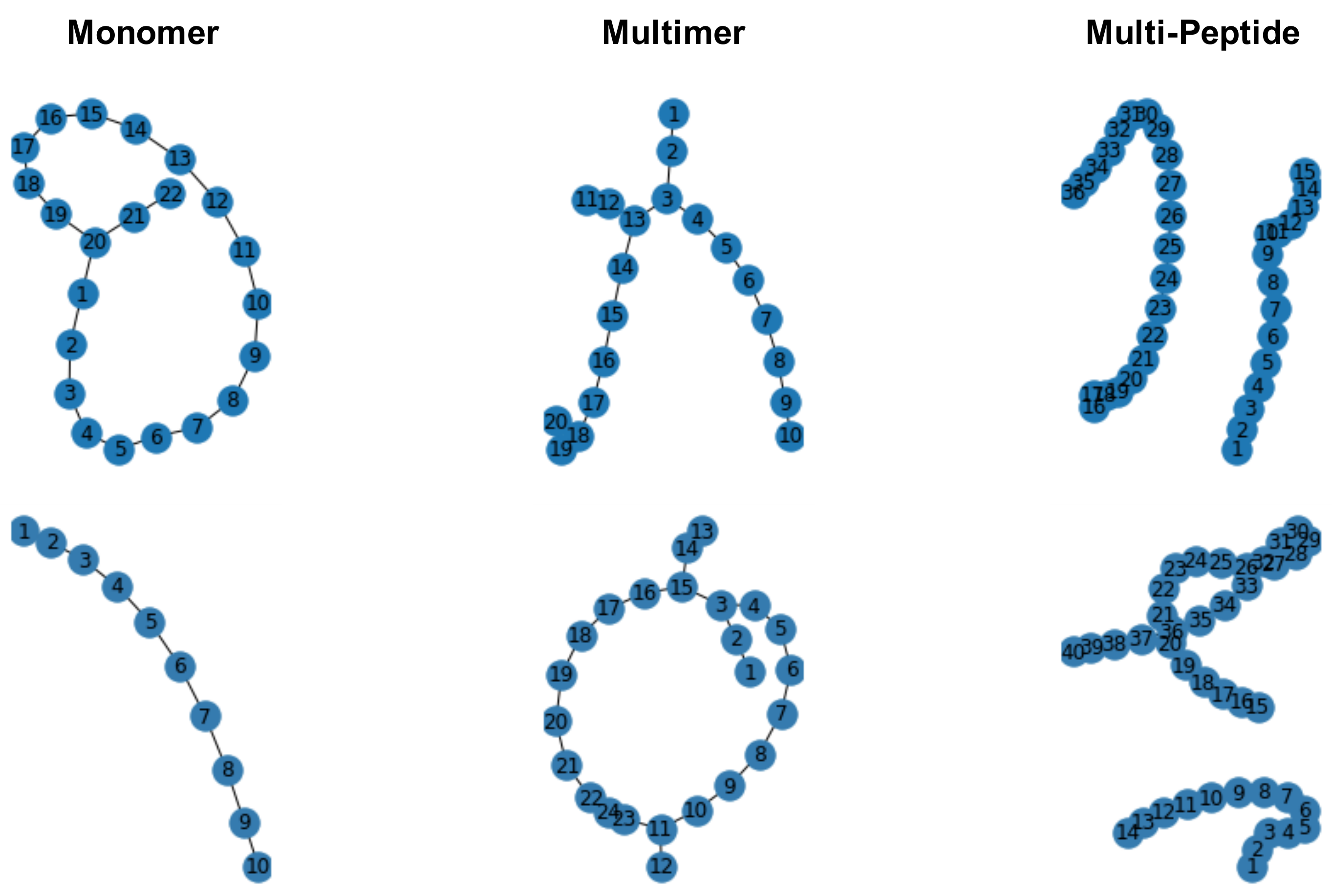}
\caption{NetworkX graph representations of peptides of all three complexities of varying molecular weights.}
\label{sifig16}
\end{figure}

A text file parser converts the macromolecule information stored in the .txt file to a NetworkX graph with monomers expressed as nodes and bonds expressed as edges. The parser goes through the .txt file line by line, stores the monomer information in a dictionary with keys as integer positions and values as monomer abbreviations, and stores the bond information in a dictionary with keys as tuples containing bond connectivities and values as bond abbreviations. Afterwards, the reader uses NetworkX to add each key in the monomer dictionary as a node and each key in the bond dictionary as an edge, storing the abbreviations as attributes for the corresponding node or edge. The resulting NetworkX graphs include both linear and highly branched architectures (Figure \ref{sifig15}, \ref{sifig16}).

\section{Node and edge attributes}
\label{si_4}
\subsection{Fingerprints – generation and optimization of hyperparameters}
\label{si_4_1}
\begin{figure}[h]
\centering
\includegraphics[width=1\textwidth]{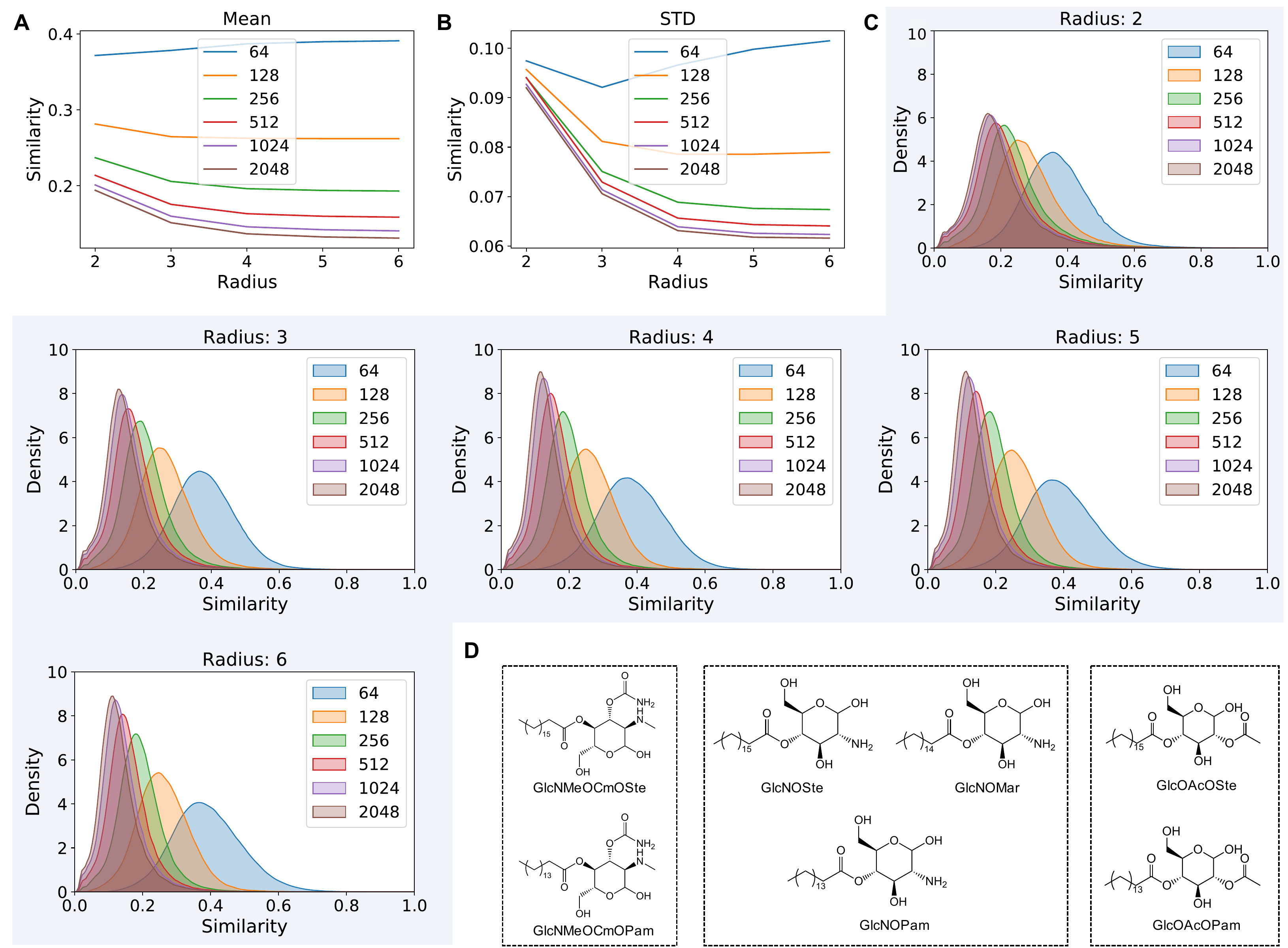}
\caption{\textbf{A.} Mean, \textbf{B.} standard deviation and \textbf{C.} distribution of Tanimoto similarity of all monomers in the glycans dataset, calculated using stereochemical extended connectivity fingerprints of different radii and bits. \textbf{D.} Glycans in the same box have the exact fingerprint for radius 3 and 128 bits.}
\label{sifig17}
\end{figure}

We used RDKit to generate stereochemical extended connectivity fingerprints \citep{Landrum2006, Rogers2010Extended}. Radius and number of bits were optimized by calculating mean and standard deviation, and visualizing the distribution of Tanimoto similarity of all monomers in the glycans dataset (Figure \ref{sifig17}A-C). We aimed to obtain fingerprints with lower number of bits and optimal radius that could represent the monomer aptly in both similarity computation and graph neural network models. For 64 bits, we observed that the mean similarity was as high as 0.4 and standard deviation for similarity went up after radius 3 indicating higher hash collision and lesser differentiability. For 128 bits, the mean and standard deviation plateaued around 0.3 and 0.08, respectively. Further, the similarity distribution was qualitatively spread over a larger range, as compared to fingerprints with larger number of bits. For fingerprint bits longer than 128, we observed lower mean and standard deviation, and decreasing spread in the similarity distribution. The decreasing trend indicates that the fingerprints with bits higher than 128 are equally dissimilar, thus, if used, will lead to glycans with equally high dissimilar scores. Hence, we chose fingerprints with 128 bits and radius 3 to generate node attributes for glycans. For edge attributes, we had 3 types of glycosidic bonds differing by the stereochemistry alone, hence, we chose fingerprints with 16 bits and radius 3. We employed similar reasoning for the peptides to select 128 bits and radius 3 for both monomers and bonds. 

We observed that there were 3 sets with 7 glycans which had exact fingerprints (Figure \ref{sifig17}D). Since, the difference was in the number of carbon atoms in the aliphatic chain, we used the fingerprints as is.

\subsection{One-hot encoding benchmark}
\label{si_4_2}
One-hot encodings of the 946 monomer and 3 bond types for glycans and the 461 monomer and 22 bond types for peptides were also employed as feature types for benchmarking with featurization using molecular fingerprints. The dimensions of the node and edge one-hot encoding features are 946 and 3, respectively, for glycans and 461 and 22, respectively, for peptides.

\section{Similarity computation}
\label{si_5}
\subsection{Analysis of graphs}
\label{si_5_1}
Most glycan graphs are sparse (Figure \ref{sifig8}). Complete graphs have density of 1, while graphs with density $\ >$ 0.5 have been defined as dense \citep{Borgwardt2020}. 

\begin{figure}[h]
\centering
\includegraphics[width=1\textwidth]{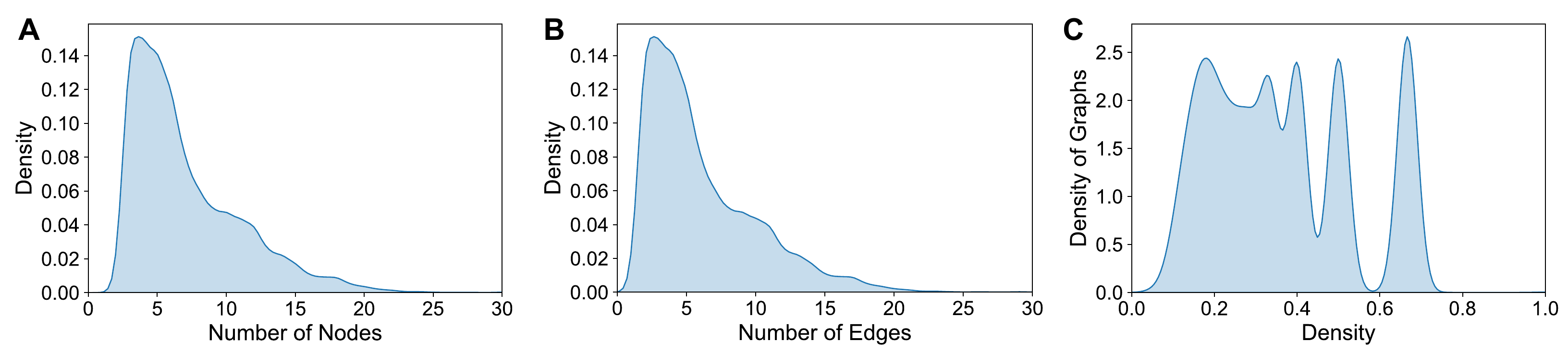}
\caption{Distribution of number of \textbf{A.} nodes, \textbf{B.} edges and \textbf{C.} density for glycan graphs.}
\label{sifig18}
\end{figure}

\subsection{Similarity matrix}
\label{a_5_2}
We computed the$\ (n \times n)$ similarity matrix for all glycans with labels on at least one taxonomic level using propagation attribute kernel in GraKeL (Figure \ref{sifig18}) \citep{Siglidis2020}. Each pair of graph similarity was computed for a maximum of 100 iterations. This resulted in 5\% of the pairs being assigned a 0 similarity (10\% of all indices in the similarity matrix are 0).

It may be noted that the computation of similarity using graph kernels is way more accessible than graph edit distances. The current computation was done on in minutes (wall time), parallelized across 24 cores. From visual inspection using htop, only about 30\% of 2 cores were being used at any particular time, and less than 5\% of all other cores were used, although there were 24 jobs running in parallel.

\begin{figure}[h]
\centering
\includegraphics[width=1\textwidth]{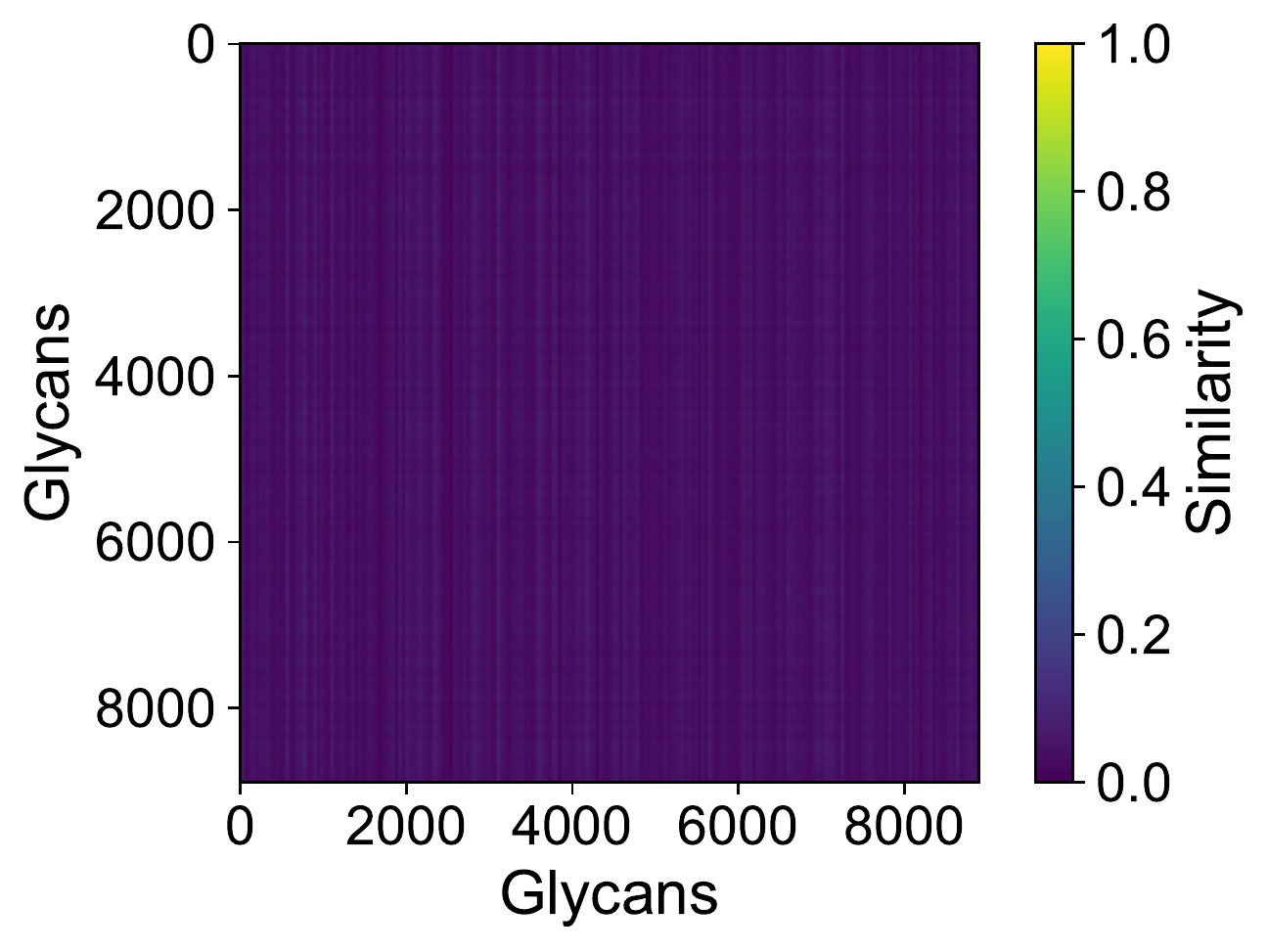}
\caption{2D plot for the$\ (n \times n)$ similarity matrix of glycans. This is not symmetric because each row is normalized by its maximum.}
\label{sifig19}
\end{figure}

\subsection{Dissimilarity computation using exact graph edit distances}
\label{a_5_3}
\begin{figure}[h]
\centering
\includegraphics[width=1\textwidth]{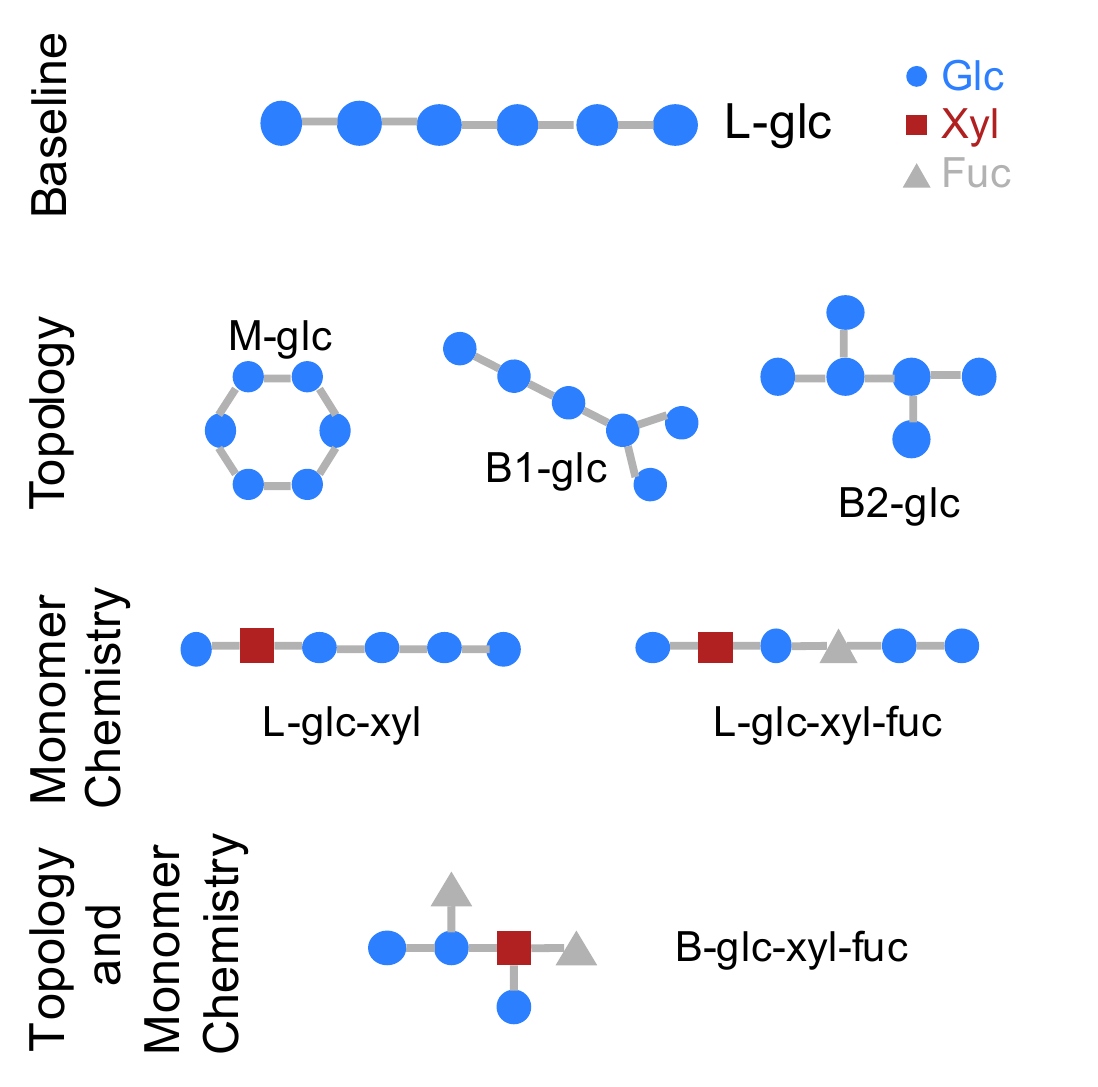}
\caption{Representative glycans used in the hyperparameter search for dissimilarity computation using exact graph edit distances, and similarity computation using graph kernels.}
\label{sifig20}
\end{figure}

We performed a grid search over combinations of possible node/edge substitution costs and multipliers for node/edge insertion/deletion costs to find an optimal set of values. Instead of similarity, we use the opposite – dissimilarity – as it is more intuitive when compared to a baseline case. We use the representative glycans in Figure 2B for this experiment, and present them here in Figure \ref{sifig20} with the names we use in Table \ref{sitab1}. 

For node/edge substitution, the Tanimoto distance between the stereochemical fingerprints, a value in the range of 0 and 1 (where 0 is a perfect match), is multiplied by the substitution cost to obtain the final graph edit distance. For node/edge insertion/deletion, a constant value which is the node/edge insertion/deletion is added to the graph edit distance. The baseline case where the edit distance is calculated for the macromolecule with itself is 0.

\begin{table}[!ht]
\centering
\caption{Dissimilarities computed using exact graph edit distances using different node/edge insertion/deletion and substitution multipliers. The remarks column notes down the hyperparameters, and baseline and other classes of glycans. Abbreviations - Hparam: hyperparameter, L: linear, B: branched, glc: glucose, xyl: xylose, fuc: fucose.}
\label{sitab1}
\resizebox{\textwidth}{!}{%
\begin{tabular}{@{}cccccccccccccccccc@{}}
\toprule
\textbf{Remarks}                   & \textbf{Hyperparameters/Glycans}                                         & \multicolumn{16}{c}{\textbf{Cases}}                                                                          \\ \midrule  
HParam1                            & \textbf{Insertion/Deletion Cost} & 1    & 1  & 1  & 1  & 3     & 3    & 3     & 3  & 5     & 5    & 5     & 5     & 10    & 10   & 10    & 10    \\ \midrule 
HParam2                            & \textbf{Substitution Cost}                                                        & 1    & 3  & 5  & 10 & 1     & 3    & 5     & 10 & 1     & 3    & 5     & 10    & 1     & 3    & 5     & 10    \\ \midrule 
Baseline                           & \textbf{L-glc}                                                                    & 0    & 0  & 0  & 0  & 0     & 0    & 0     & 0  & 0     & 0    & 0     & 0     & 0     & 0    & 0     & 0     \\ \midrule 
\multirow{3}{*}{Topology}          & \textbf{M-glc}                                                                    & 1    & 1  & 1  & 1  & 3     & 3    & 3     & 3  & 5     & 5    & 5     & 5     & 10    & 10   & 10    & 10    \\ \cmidrule(l){2-18} 
                                   & \textbf{B1-glc}                                                                   & 2    & 2  & 2  & 2  & 6     & 6    & 6     & 6  & 10    & 10   & 10    & 10    & 20    & 20   & 20    & 20    \\ \cmidrule(l){2-18} 
                                   & \textbf{B2-glc}                                                                   & 4    & 4  & 4  & 4  & 12    & 12   & 12    & 12 & 20    & 20   & 20    & 20    & 40    & 40   & 40    & 40    \\ \midrule 
\multirow{2}{*}{Monomer Chemistry} & \textbf{L-glc-xyl}                                                                & 0.68 & 6  & 6  & 6  & 0.68  & 2.05 & 3.42  & 18 & 0.68  & 2.05 & 3.42  & 6.84  & 0.68  & 2.05 & 3.42  & 6.84  \\ \cmidrule(l){2-18} 
                                   & \textbf{L-glc-xyl-fuc}                                                            & 1.43 & 12 & 12 & 12 & 1.43  & 4.28 & 7.13  & 36 & 1.43  & 4.28 & 7.13  & 14.26 & 1.43  & 4.28 & 7.13  & 14.26 \\ \midrule 
Topology and Chemistry             & \textbf{B-glc-xyl-fuc}                                                            & 6.17 & 14 & 14 & 14 & 14.17 & 18.5 & 22.84 & 42 & 22.17 & 26.5 & 30.84 & 41.67 & 42.17 & 46.5 & 50.84 & 61.67 \\ \bottomrule
\end{tabular}%
}
\end{table}

\subsection{Similarity computation using graph kernels}
\label{a_5_4}
We performed a grid search over hyperparameters of propagation attribute kernel – bin width {1, 3, 10, 100}, and the preserved distance metric on local sensitive hashing {‘L1-norm’, ‘L2-norm’}. We used 30 as the maximum number of iterations for the kernel computation for this experiment. We use the representative glycans Figure \ref{sifig20}, and present the results in Table \ref{sitab2}. 

\begin{table}[!ht]
\centering
\caption{Similarities computed using graph kernels with different values of bin width and distance metric. The remarks column notes down the hyperparameters, and baseline and other classes of glycans. Abbreviations - Hparam: hyperparameter, L: linear, B: branched, glc: glucose, xyl: xylose, fuc: fucose.}
\label{sitab2}
\resizebox{\textwidth}{!}{%
\begin{tabular}{@{}cccccccccc@{}}
\toprule
\textbf{Remarks}                   & \textbf{Hyperparameters/Glycans} & \multicolumn{8}{c}{\textbf{Cases}}           \\ \midrule
HParam1                            & \textbf{Bin width}                        & 1   & 1   & 3   & 3   & 10  & 10  & 100 & 100 \\ \midrule
HParam2                            & \textbf{Distance metric}                  & L1  & L2  & L1  & L2  & L1  & L2  & L1  & L2  \\ \midrule
Baseline                           & \textbf{L-glc}                            & 230 & 230 & 230 & 230 & 230 & 230 & 230 & 230 \\ \midrule
\multirow{3}{*}{Topology}          & \textbf{M-glc}                            & 61  & 61  & 61  & 61  & 61  & 61  & 61  & 61  \\ \cmidrule{2-10} 
                                   & \textbf{B1-glc}                           & 56  & 56  & 56  & 56  & 56  & 56  & 56  & 56  \\ \cmidrule{2-10} 
                                   & \textbf{B2-glc}                           & 51  & 51  & 51  & 51  & 51  & 51  & 51  & 51  \\ \midrule
\multirow{2}{*}{Monomer Chemistry} & \textbf{L-glc-xyl}                       & 50  & 50  & 50  & 50  & 50  & 55  & 56  & 61  \\ \cmidrule{2-10} 
                                   & \textbf{L-glc-xyl-fuc}                    & 39  & 39  & 39  & 39  & 45  & 44  & 51  & 61  \\ \midrule
Topology and Chemistry             & \textbf{B-glc-xyl-fuc}                    & 23  & 23  & 23  & 23  & 35  & 33  & 46  & 51  \\ \bottomrule
\end{tabular}%
}
\end{table}

\section{Dimensionality reduction}
\label{si_6}
\subsection{Hyperparameter optimization for Uniform Manifold Approximation and Projection (UMAP)}
\label{si_6_1}
\begin{figure}[h]
\centering
\includegraphics[width=1\textwidth]{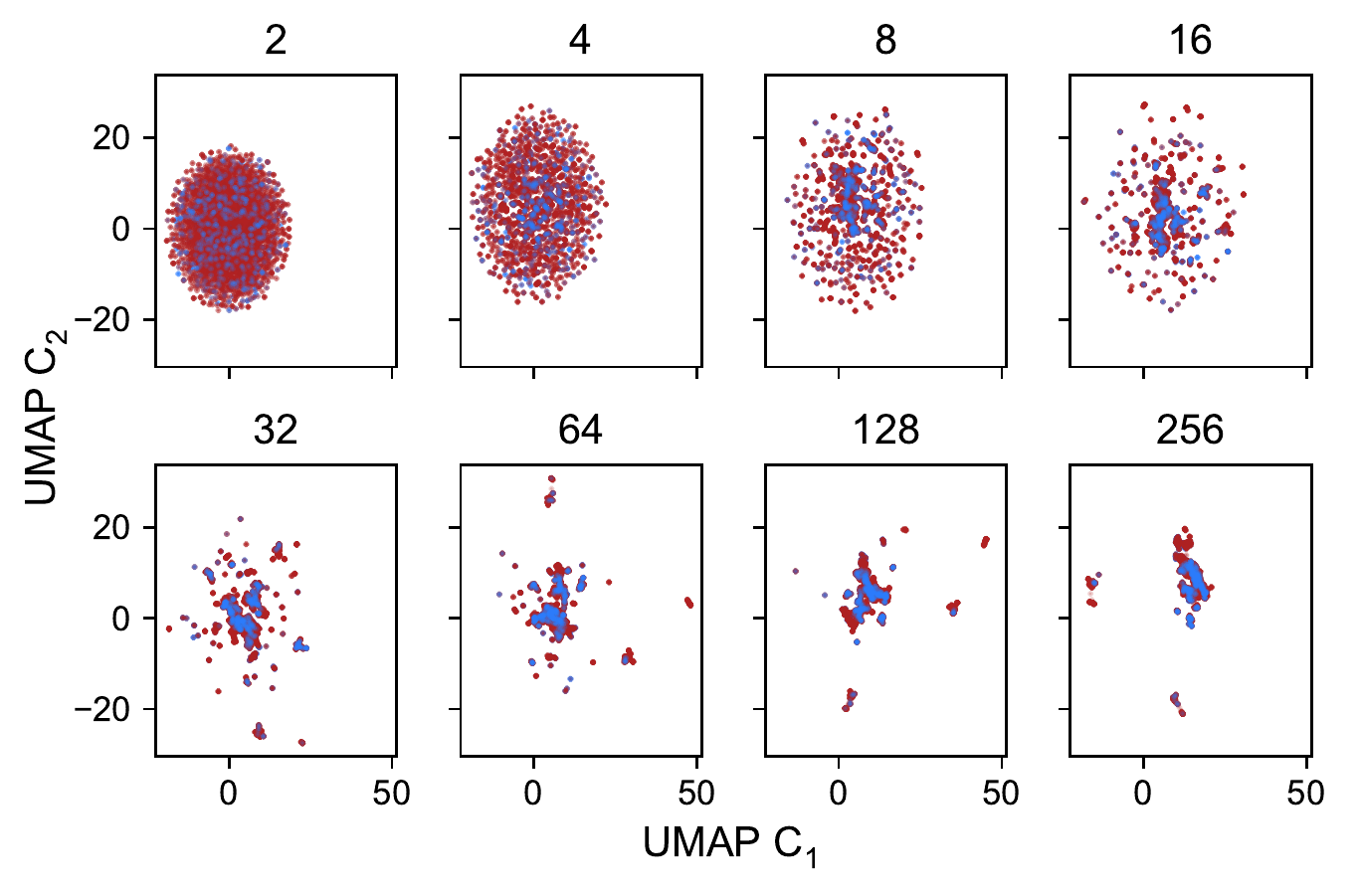}
\caption{Visualization of scatter plots for UMAP components, obtained by dimensionality reduction of similarity vectors. The number of neighbors for each UMAP computation has been noted in the title of the respective sub-plot. Coloration is by immunogenicity (red: immunogenic, blue: non-immunogenic glycans).}
\label{sifig21}
\end{figure}

Number of neighbors was optimized for 2-component UMAP dimensionality reduction of similarity vectors (Figure \ref{sifig21}) \citep{McInnes2018}. From visual inspection, UMAP with 128 neighbors seems to resolve into optimal size and number of clusters. The subplot shows distinct regions for the immunogenic and non-immunogenic glycans. We note that there is more to similarity than graph kernel distances, and observe that it has been effectively captured using the GNN models (Appendix \ref{si_7}).

\begin{figure}[h]
\centering
\includegraphics[width=1\textwidth]{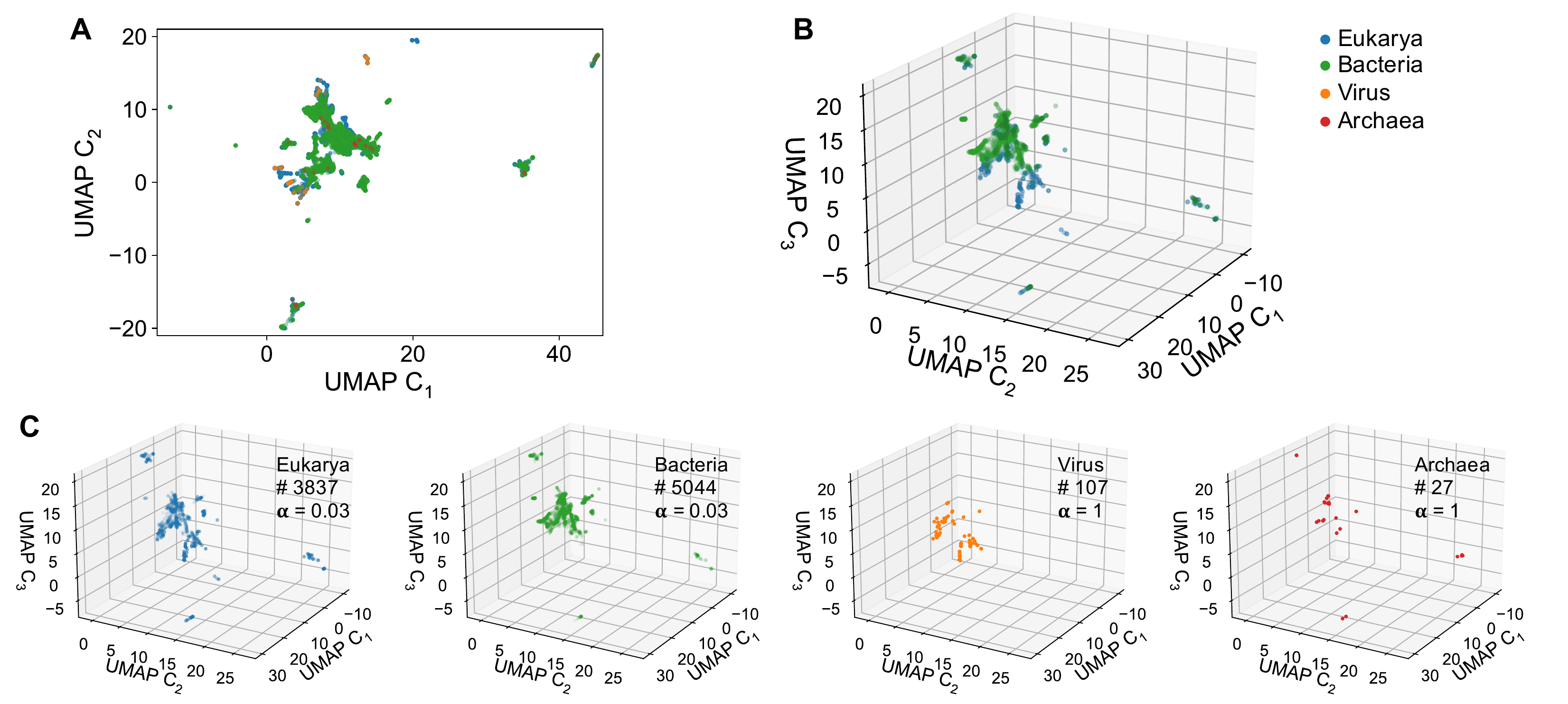}
\caption{Visualization of \textbf{A.} 2 and \textbf{B.} 3 components UMAP, colored by domain. \textbf{C.} Glycans corresponding to individual domains are shown, with the text noting the domain, number of glycans, and transparency ($\alpha$) of each point on a scale of 0 to 1, where 1 is opaque}
\label{sifig22}
\end{figure}

\begin{figure}[h]
\centering
\includegraphics[width=0.5\textwidth]{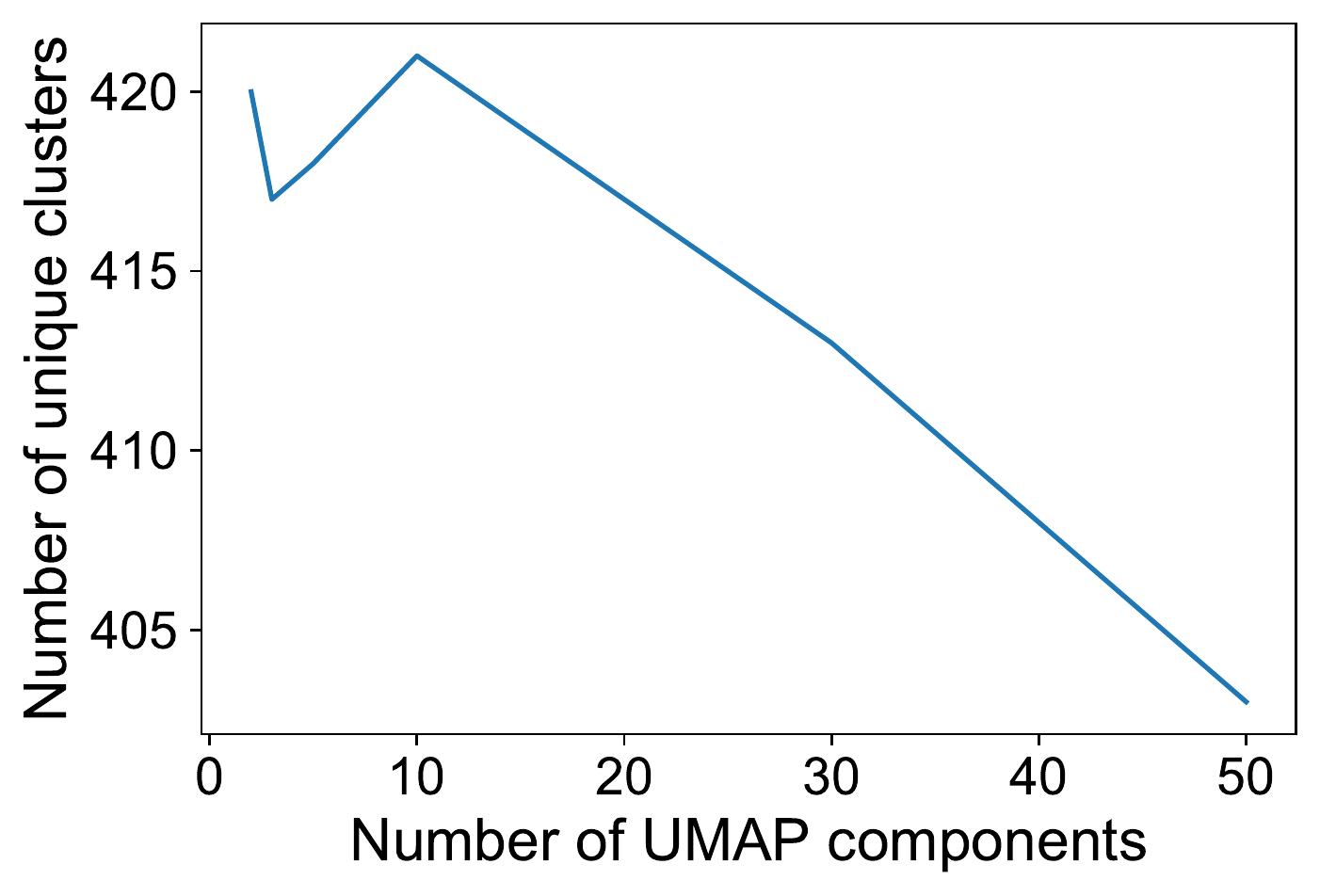}
\caption{Number of HDBSCAN unsupervised clusters obtained from UMAP dimensionality reduction for different components.}
\label{sifig23}
\end{figure}

\begin{figure}[h]
\centering
\includegraphics[width=1\textwidth]{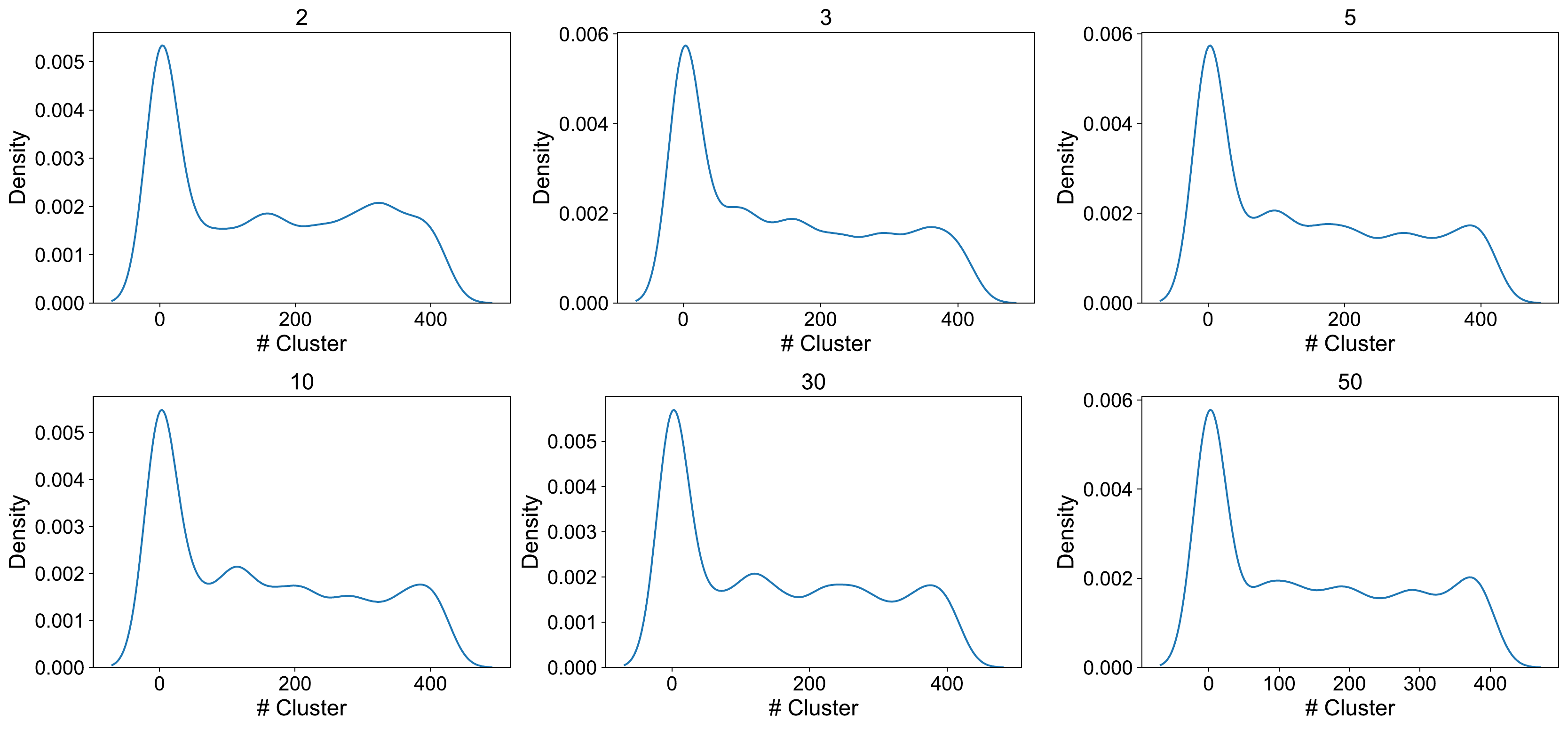}
\caption{Distribution of glycans in each cluster for HDBSCAN unsupervised clusters obtained from UMAP dimensionality reduction for different components. The components have been noted as titles for the sub-plots.}
\label{sifig24}
\end{figure}

UMAP dimensionality reduction for higher number of components does not provide more information. The 3 components UMAP looked similar to the 2 components, with slightly more disentangled families (Figure \ref{sifig22}). We limited our visual analysis to 3 components. To check if more components can help in finding distinct clusters, we did dimensionality reduction for $\{2, 3, 5, 10, 30, 50\}$ components and let HDBSCAN - an unsupervised clustering algorithm – to figure out how many clusters are there \citep{McInnes2017}. We noted that the number of clusters are pretty similar, and in low 400s (Figure \ref{sifig23}). The high number of clusters indicates the diversity of the space, and the differences in terms of taxonomy. As a further check to see if the distribution of glycans in different clusters is different, we plotted the histograms of the glycans assigned to each cluster. Across all the components, the histograms seem to be consistent with the number of glycans in each of them (Figure \ref{sifig24}).

\subsection{T-SNE benchmark}
\label{si_6_2}
\begin{figure}[h]
\centering
\includegraphics[width=0.5\textwidth]{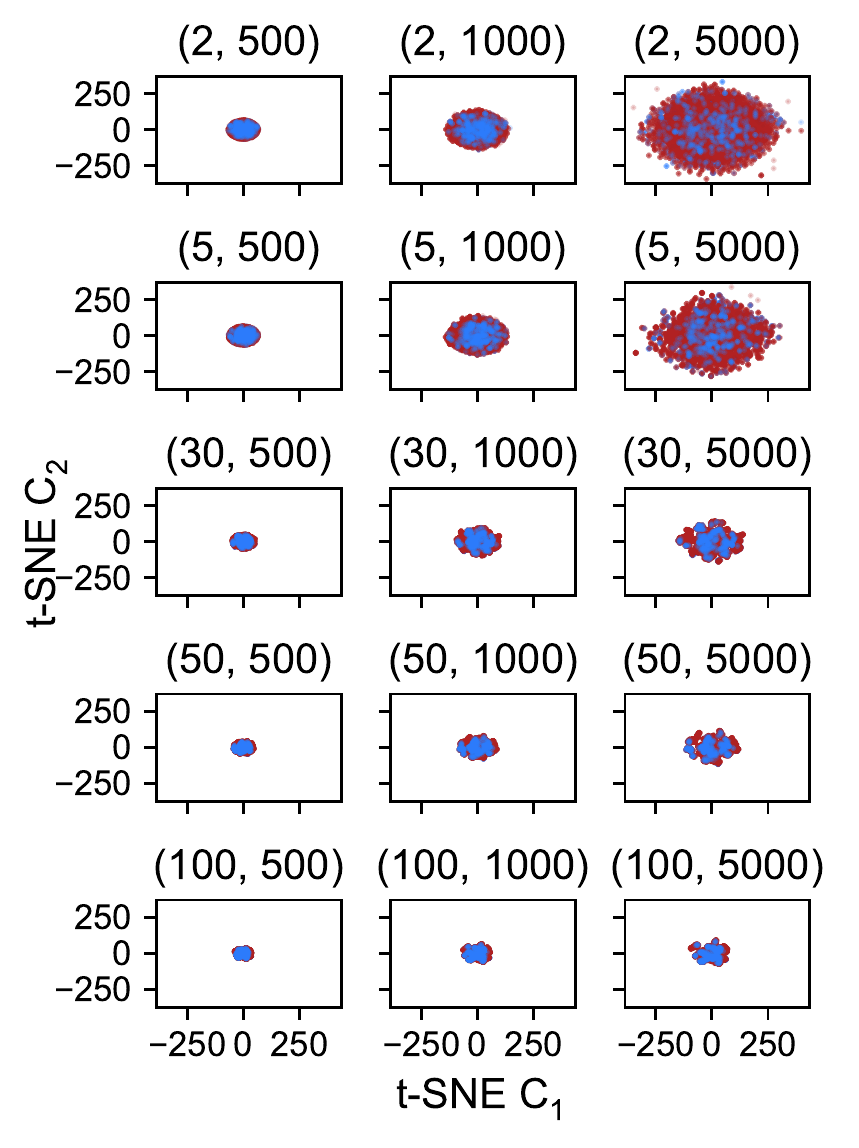}
\caption{Visualization of scatter plots for t-SNE components, obtained by dimensionality reduction of similarity vectors. The (perplexity, number of steps) for each t-SNE computation has been noted in the title of the respective sub-plot. Coloration is by immunogenicity (red: immunogenic, blue: non-immunogenic glycans).}
\label{sifig25}
\end{figure}

We benchmarked the dimensionality reduction results obtained from UMAP against a broad range of t-stochastic neighbor embeddings (t-SNE) models \citep{VanDerMaaten2008VisualizingT-SNE}. For the different models, we varied perplexity as $\{2, 5, 30, 50, 100\}$, and number of steps as $\{500, 1000, 5000\}$. From the scatter plot, colored by immunogenicity labels, we noted that dimensionality reduction using t-SNE was not able to deduce the differences and getting the glycans into distinct areas (Figure \ref{sifig25}).

\section{Supervised learning with graph neural networks}
\label{si_7}
\subsection{Deep graphs library (DGL) graphs}
\label{si_7_1}
Following featurization, NetworkX graphs were converted into undirected, unweighted, and homogenous DGL graphs \citep{Wang2019c}. For GCN and GAT model architectures, self-loops were added to the DGL graphs to prevent silent performance regression due to zero-in-degree nodes during training. 

\subsection{Model architectures}
\label{si_7_2}
We performed graph classification using five distinct graph neural network model architectures detailed below:
\begin{itemize}
    \item Weave \citep{Kearnes2016}
    \item Message Passing Neural Networks (MPNNs) \citep{Gilmer2017}
    \item Attentive FP \citep{Xiong2020}
\item Graph convolutional networks (GCN) \citep{Kipf2019}
\item Graph Attention Networks (GAT) \citep{Romero2018}
\end{itemize}

While Weave, MPNN, and Attentive FP utilize both node and edge attributes in prediction, GCN and GAT only consider node attributes. The models were trained using implementations in the DGL LifeSci library \citep{Wang2019c}. For classification, the optimization was done by minimization of average cross-entropy loss between batches and additional metrics such as F1 score, recall, precision and accuracy were noted. For regression, the optimization was done by minimization of root-mean-squared-error loss on the validation dataset. 

\subsection{Glycan graphs classification}
\label{si_7_3}
\subsubsection{Immunogenicity}
\label{si_7_3_1}
\textbf{Dataset.} 1313 glycans in the database have immunogenicity labels, 631 of which are immunogenic and 682 of which are not immunogenic. The training was performed on 60\%, validated on 20\%, and tested on held-out 20\% data. 

\begin{figure}[h]
\centering
\includegraphics[width=1\textwidth]{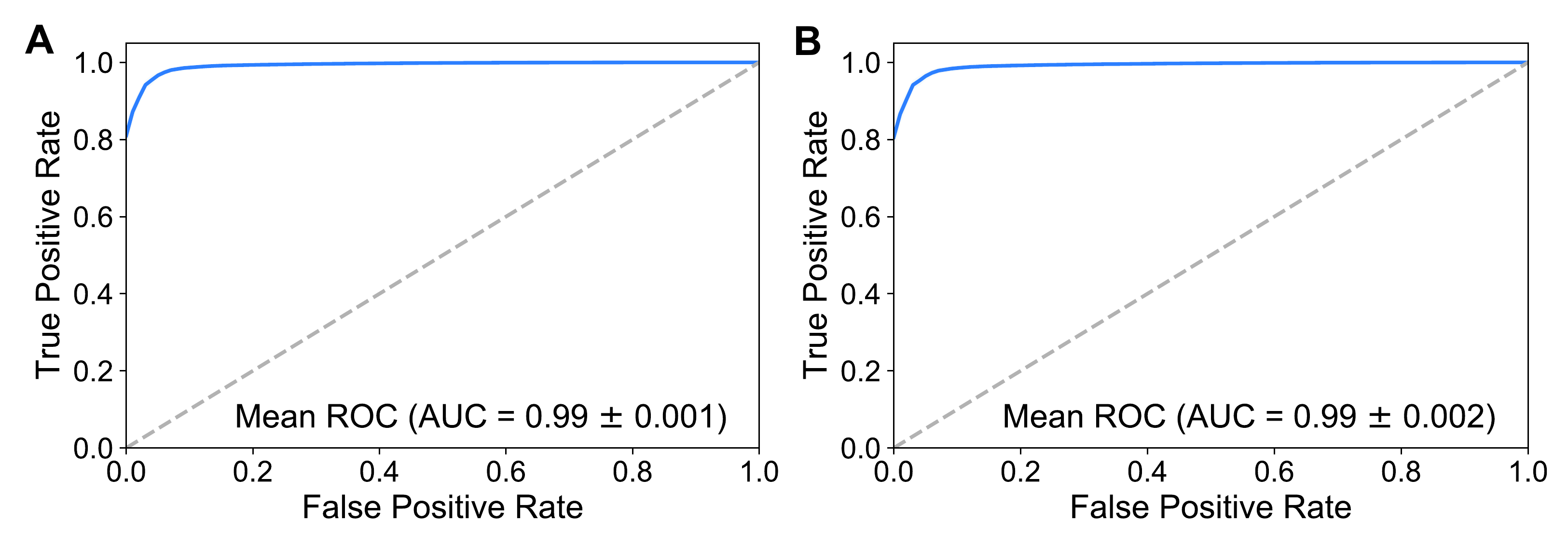}
\caption{ROC-AUC curves for fingerprint and one-hot encoding-featurized graphs. \textbf{A.} Mean ROC-AUC curve of all 25 fingerprint-featurized experiments (5 model architectures with 5 sets of hyperparameters for each architecture), with the standard deviation shaded in light blue too insignificant to be visible in the graph. \textbf{B.} Mean ROC-AUC curve of all 25 one-hot encoding-featurized experiments.}
\label{sifig26}
\end{figure}

\begin{figure}[h]
\centering
\includegraphics[width=1\textwidth]{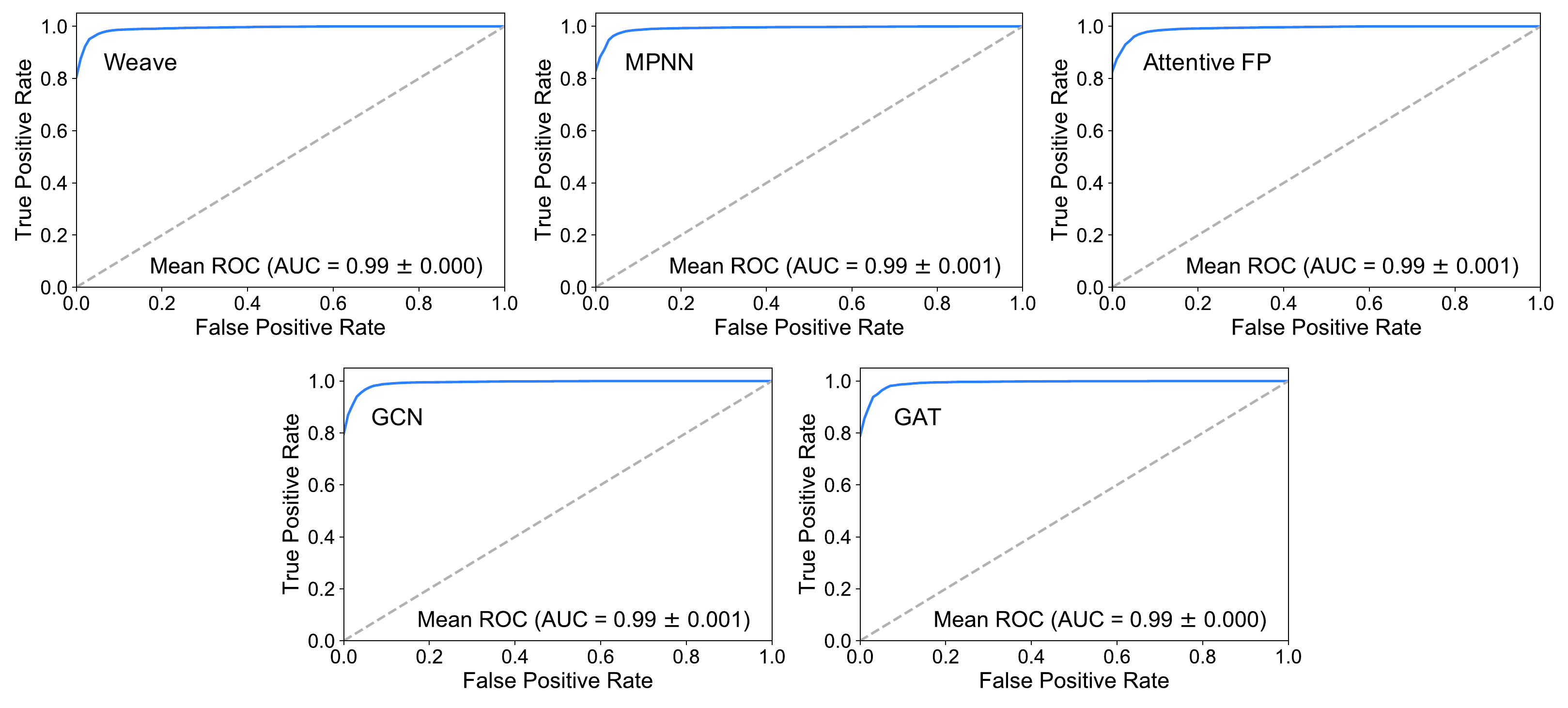}
\caption{Mean ROC-AUC curves for fingerprint-featurized experiments for each of the five model architectures (Weave, MPNN, Attentive FP, GCN, and GAT), with the standard deviation shaded in light blue too insignificant to be visible. A standard deviation of 0.000 denotes a value of$\ <$0.001.}
\label{sifig27}
\end{figure}

\begin{figure}[h]
\centering
\includegraphics[width=1\textwidth]{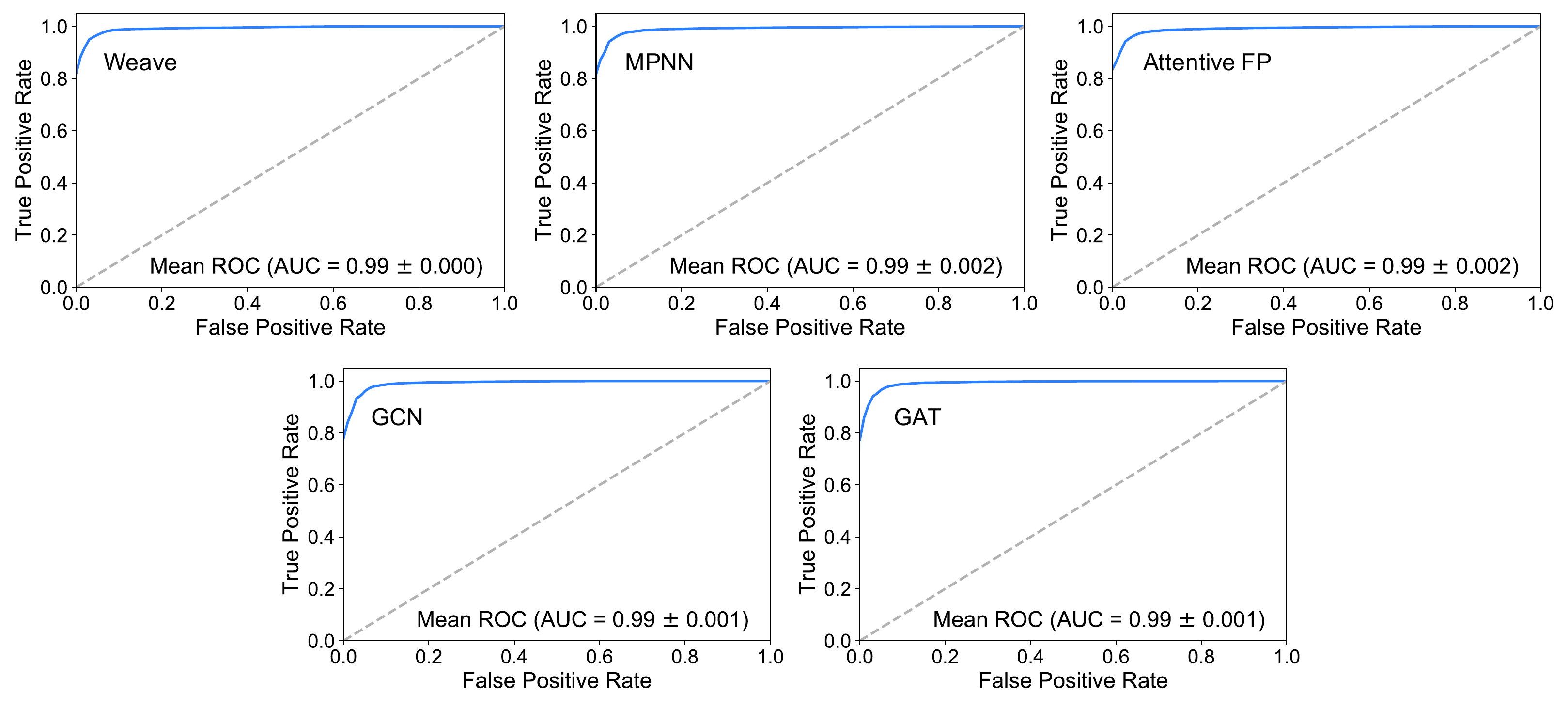}
\caption{Mean ROC-AUC curves for one-hot encoding-featurized experiments for each of the five model architectures (Weave, MPNN, Attentive FP, GCN, and GAT), with the standard deviation shaded in light blue too insignificant to be visible. A standard deviation of 0.000 denotes a value of$\ <$0.001.}
\label{sifig28}
\end{figure}

\textbf{Models.} We classified immunogenicity using 5 model architectures combined with 2 different node and edge featurization types, for a total of 10 model architecture-attribute pairs. For each benchmark, hyperparameter optimization against minimization of binary cross entropy loss was performed on SigOpt, a standardized hyperparameter optimization platform, for 1000 observations and the 5 best sets of hyperparameters were extracted. Each model architecture and featurization combination was trained using the 5 best sets of hyperparameters from SigOpt using 5 distinct random seeds for splitting the dataset into train-validation-test datasets, for a total of 25 trainings per model per attribute type. The tables below report the metrics for the most optimal set of hyperparameters, the values of the most optimal hyperparameters, and mean metrics for each training across all 25 runs. All models achieve stellar performance on all metrics, with little meaningful difference in performance between fingerprint and one-hot encoding featurization (Figures \ref{sifig26}-\ref{sifig28} ; Table \ref{sitab3}).

\begin{table}[t]
\centering
\caption{Test dataset metrics for the most optimal set of hyperparameters, or the set of hyperparameters that results in the lowest loss. For each metric, the mean $\mu$ and standard deviation $\sigma$ are displayed across all 5 random seeds. “FP” denotes condensed fingerprint featurization, and “One-hot” denotes one-hot encoding featurization.}
\label{sitab3}
\resizebox{\textwidth}{!}{%
\begin{tabular}{@{}cccccccccccccc@{}}
\toprule
Model &
  Attribute   Type &
  \multicolumn{2}{c}{ROC-AUC} &
  \multicolumn{2}{c}{F1} &
  \multicolumn{2}{c}{Recall} &
  \multicolumn{2}{c}{Precision} &
  \multicolumn{2}{c}{Accuracy} &
  \multicolumn{2}{c}{Loss} \\ \midrule
                             &         & $\mu$    & $\sigma$    & $\mu$    & $\sigma$    & $\mu$    & $\sigma$    & $\mu$    & $\sigma$    & $\mu$    & $\sigma$    & $\mu$    & $\sigma$    \\ \midrule
\multirow{2}{*}{Weave}       & FP      & 0.990 & 0.005 & 0.956 & 0.011 & 0.955 & 0.012 & 0.957 & 0.011 & 0.956 & 0.011 & 0.133 & 0.106 \\ \cmidrule(l){2-14} 
                             & One-hot & 0.992 & 0.005 & 0.962 & 0.007 & 0.962 & 0.007 & 0.962 & 0.007 & 0.962 & 0.007 & 0.105 & 0.055 \\ \midrule
\multirow{2}{*}{MPNN}        & FP      & 0.985 & 0.009 & 0.955 & 0.012 & 0.955 & 0.012 & 0.957 & 0.012 & 0.955 & 0.012 & 0.137 & 0.100 \\ \cmidrule(l){2-14} 
                             & One-hot & 0.988 & 0.006 & 0.953 & 0.010 & 0.953 & 0.010 & 0.954 & 0.010 & 0.953 & 0.010 & 0.123 & 0.052 \\ \midrule
\multirow{2}{*}{AttentiveFP} & FP      & 0.990 & 0.005 & 0.954 & 0.011 & 0.953 & 0.011 & 0.955 & 0.012 & 0.954 & 0.011 & 0.125 & 0.104 \\ \cmidrule(l){2-14} 
                             & One-hot & 0.990 & 0.005 & 0.951 & 0.011 & 0.951 & 0.012 & 0.952 & 0.011 & 0.952 & 0.011 & 0.120 & 0.084 \\ \midrule
\multirow{2}{*}{GCN}         & FP      & 0.992 & 0.004 & 0.953 & 0.013 & 0.953 & 0.013 & 0.954 & 0.012 & 0.953 & 0.012 & 0.110 & 0.086 \\ \cmidrule(l){2-14} 
                             & One-hot & 0.990 & 0.006 & 0.955 & 0.012 & 0.956 & 0.012 & 0.956 & 0.013 & 0.956 & 0.012 & 0.123 & 0.110 \\ \midrule
\multirow{2}{*}{GAT}         & FP      & 0.991 & 0.006 & 0.954 & 0.014 & 0.954 & 0.013 & 0.954 & 0.014 & 0.954 & 0.014 & 0.109 & 0.081 \\ \cmidrule(l){2-14} 
                             & One-hot & 0.991 & 0.004 & 0.954 & 0.011 & 0.954 & 0.011 & 0.955 & 0.010 & 0.955 & 0.011 & 0.119 & 0.085 \\ \bottomrule
\end{tabular}%
}
\end{table}

\subsubsection{Taxonomy}
\label{si_7_3_2}
\textbf{Dataset.} The taxonomy of the glycans was considered on eight levels: domain, kingdom, phylum, class, order, genus, and species. First, the classification of each species into the other seven taxonomic levels was obtained from the supplementary tables of SweetOrigins \citep{Bojar2021}. For each glycan with species information, taxonomic information was added in a new .csv file, with multiple labels on the same taxonomic level separated by commas. Afterwards, any labels with fewer than five glycans were removed, and any species names ending with “sp” that are therefore genus labels in disguise were filtered out to produce the final dataset for taxonomic training. The final dataset consists of 8899 unique glycans with labels on at least one taxonomic level encompassing 4 domains, 9 kingdoms, 33 phyla, 70 classes, 144 orders, 253 families, 400 genus, and 567 species. The training was performed on 60\%, validated on 20\%, and tested on held-out 20\% data.

Graph labels were generated as one-hot encodings for each taxonomic level, with the length of each one-hot encoding tensor corresponding with the number of unique labels in the taxonomic level. The taxonomy labels accommodate cases in which glycans possess multiple labels within the same taxonomic level. 

\begin{figure}[H]
\centering
\includegraphics[width=1\textwidth]{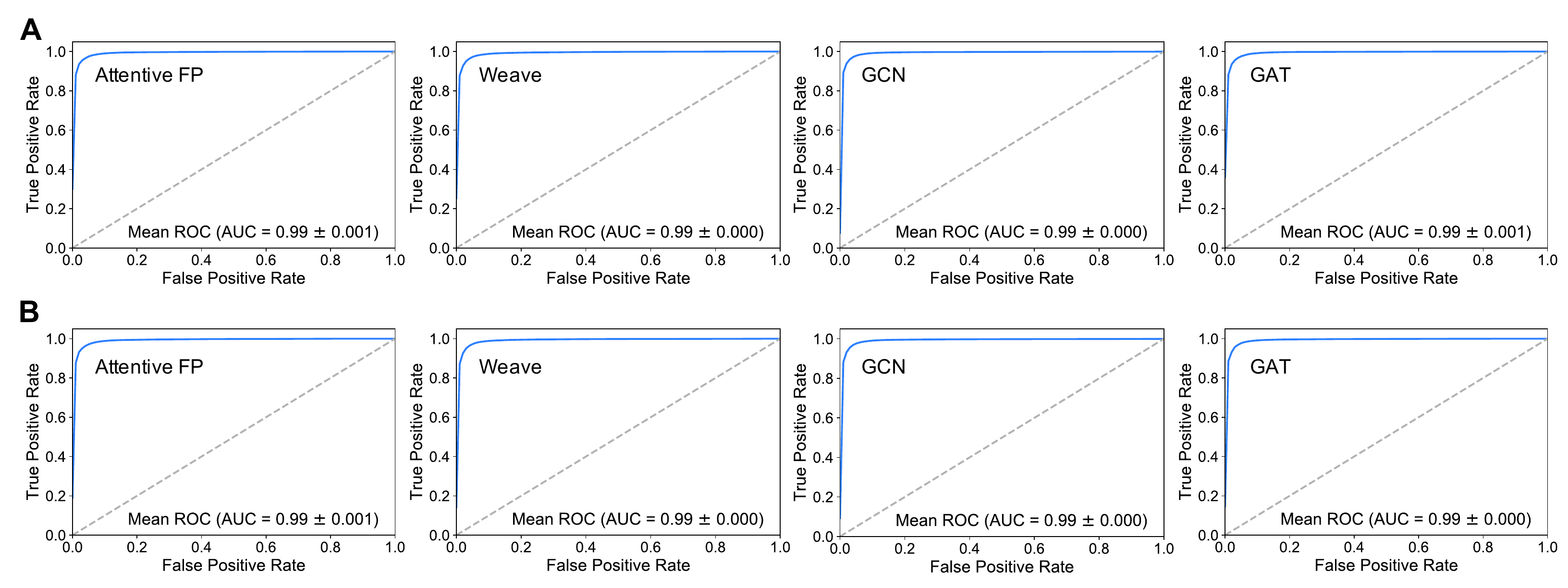}
\caption{ROC-AUC curves for classification on the domain level for each of four model architectures (Weave, Attentive FP, GCN, and GAT), with the standard deviation shaded in light blue too insignificant to be visible. A standard deviation of 0.000 denotes a value of $\ <$0.001. \textbf{A.} Mean ROC-AUC curves for fingerprint-featurized experiments, with each graph displaying the mean and standard deviation across 5 sets of hyperparameters. \textbf{B.} Mean ROC-AUC curves for one-hot encoding-featurized experiments.}
\label{sifig29}
\end{figure}

\begin{figure}[H]
\centering
\includegraphics[width=1\textwidth]{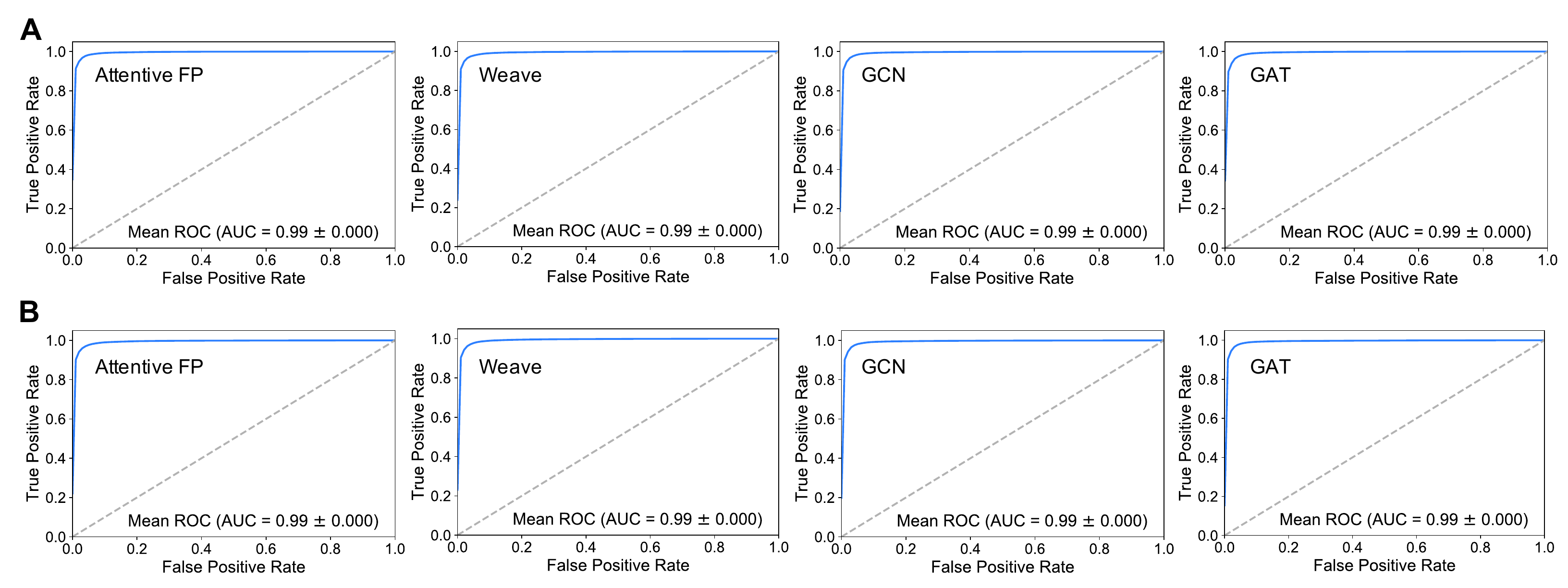}
\caption{ROC-AUC curves for classification on the kingdom level for each of four model architectures (Weave, Attentive FP, GCN, and GAT), with the standard deviation shaded in light blue too insignificant to be visible. A standard deviation of 0.000 denotes a value of $\ <$0.001. \textbf{A.} Mean ROC-AUC curves for fingerprint-featurized experiments, with each graph displaying the mean and standard deviation across 5 sets of hyperparameters. \textbf{B.} Mean ROC-AUC curves for one-hot encoding-featurized experiments.}
\label{sifig30}
\end{figure}

\begin{figure}[H]
\centering
\includegraphics[width=1\textwidth]{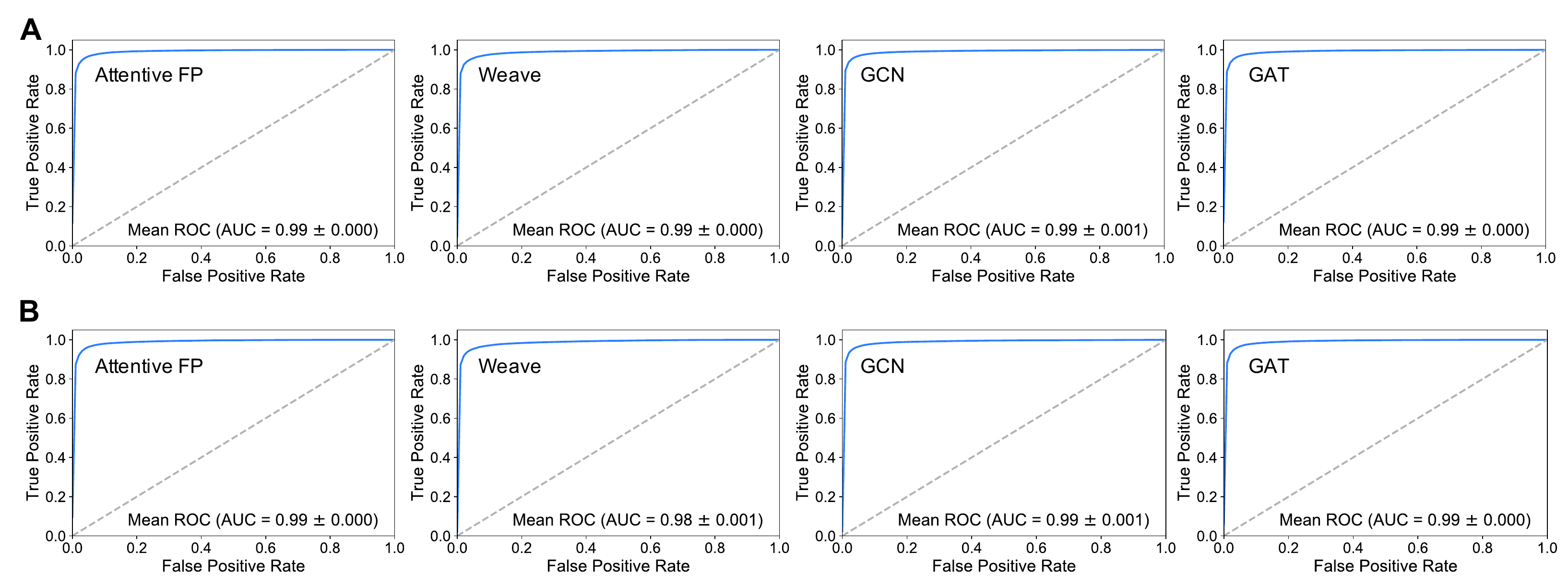}
\caption{ROC-AUC curves for classification on the phylum level for each of four model architectures (Weave, Attentive FP, GCN, and GAT), with the standard deviation shaded in light blue too insignificant to be visible. A standard deviation of 0.000 denotes a value of $\ <$0.001. \textbf{A.} Mean ROC-AUC curves for fingerprint-featurized experiments, with each graph displaying the mean and standard deviation across 5 sets of hyperparameters. \textbf{B.} Mean ROC-AUC curves for one-hot encoding-featurized experiments.}
\label{sifig31}
\end{figure}

\begin{figure}[H]
\centering
\includegraphics[width=1\textwidth]{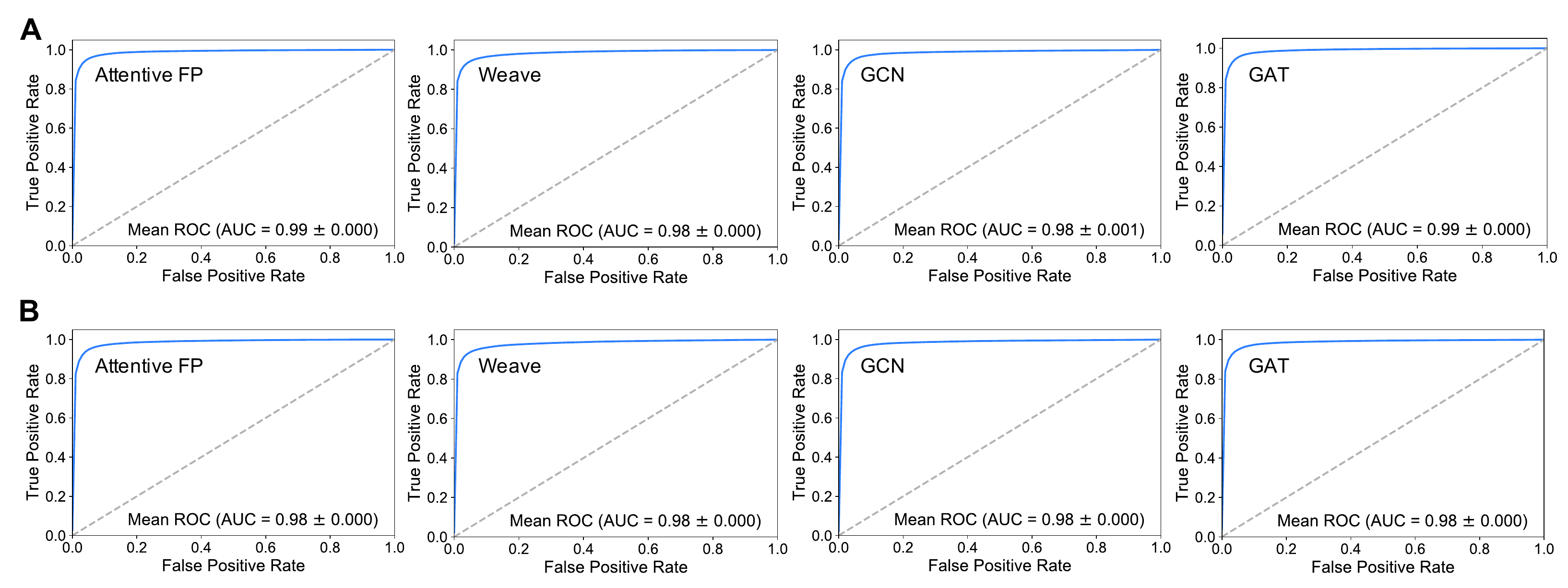}
\caption{ROC-AUC curves for classification on the class level for each of four model architectures (Weave, Attentive FP, GCN, and GAT), with the standard deviation shaded in light blue too insignificant to be visible. A standard deviation of 0.000 denotes a value of $\ <$0.001. \textbf{A.} Mean ROC-AUC curves for fingerprint-featurized experiments, with each graph displaying the mean and standard deviation across 5 sets of hyperparameters. \textbf{B.} Mean ROC-AUC curves for one-hot encoding-featurized experiments.}
\label{sifig32}
\end{figure}

\begin{figure}[H]
\centering
\includegraphics[width=1\textwidth]{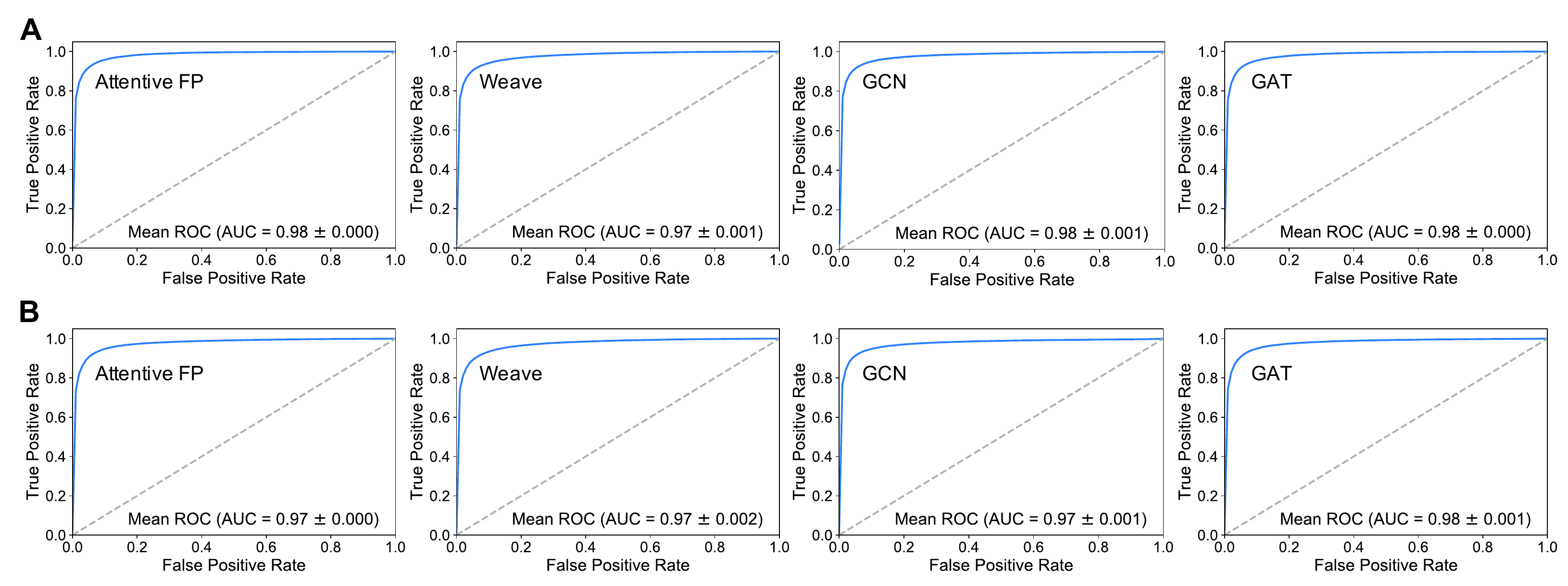}
\caption{ROC-AUC curves for classification on the order level for each of four model architectures (Weave, Attentive FP, GCN, and GAT), with the standard deviation shaded in light blue too insignificant to be visible. A standard deviation of 0.000 denotes a value of $\ <$0.001. \textbf{A.} Mean ROC-AUC curves for fingerprint-featurized experiments, with each graph displaying the mean and standard deviation across 5 sets of hyperparameters. \textbf{B.} Mean ROC-AUC curves for one-hot encoding-featurized experiments.}
\label{sifig33}
\end{figure}

\begin{figure}[H]
\centering
\includegraphics[width=1\textwidth]{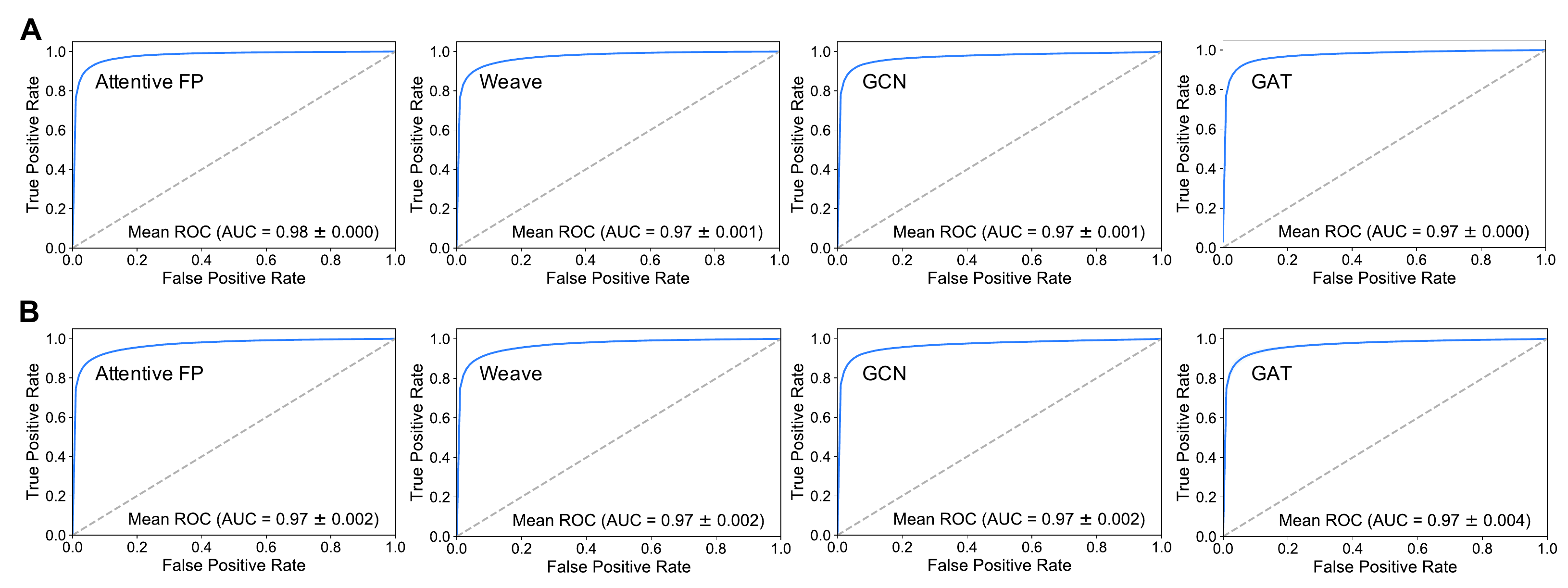}
\caption{ROC-AUC curves for classification on the family level for each of four model architectures (Weave, Attentive FP, GCN, and GAT), with the standard deviation shaded in light blue too insignificant to be visible. A standard deviation of 0.000 denotes a value of $\ <$0.001. \textbf{A.} Mean ROC-AUC curves for fingerprint-featurized experiments, with each graph displaying the mean and standard deviation across 5 sets of hyperparameters. \textbf{B.} Mean ROC-AUC curves for one-hot encoding-featurized experiments.}
\label{sifig34}
\end{figure}

\begin{figure}[H]
\centering
\includegraphics[width=1\textwidth]{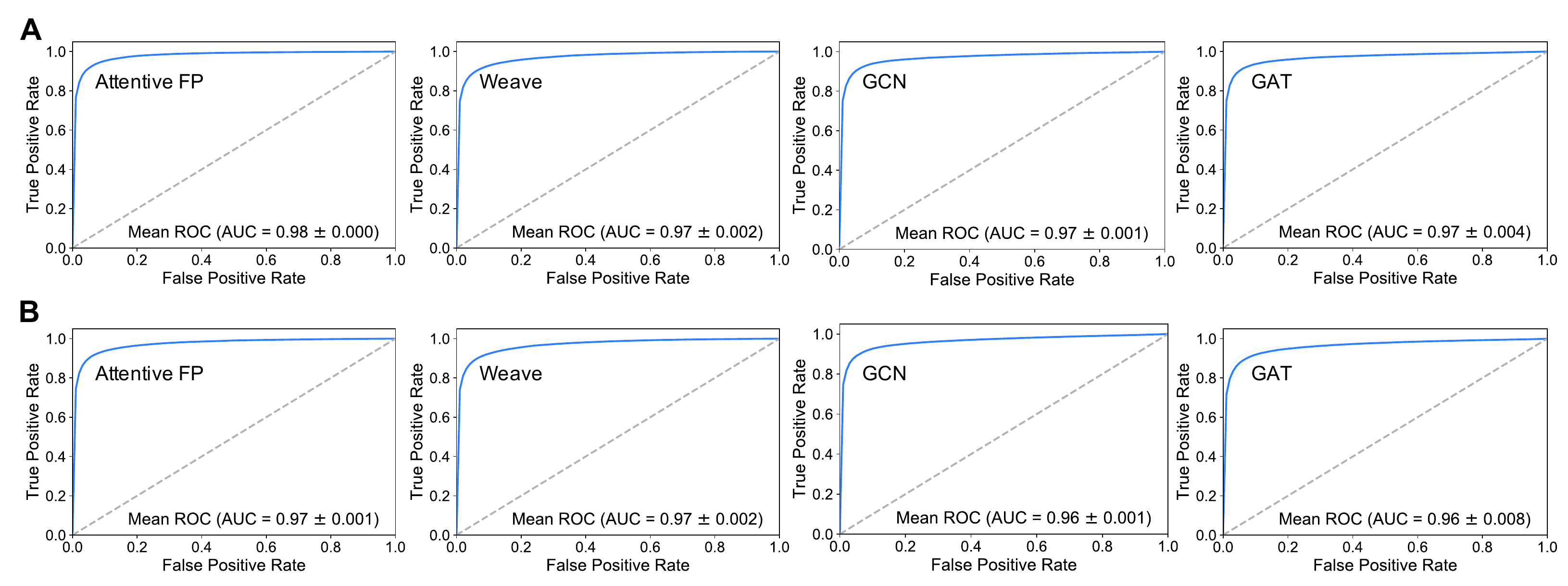}
\caption{ROC-AUC curves for classification on the genus level for each of four model architectures (Weave, Attentive FP, GCN, and GAT), with the standard deviation shaded in light blue too insignificant to be visible. A standard deviation of 0.000 denotes a value of $\ <$0.001. \textbf{A.} Mean ROC-AUC curves for fingerprint-featurized experiments, with each graph displaying the mean and standard deviation across 5 sets of hyperparameters. \textbf{B.} Mean ROC-AUC curves for one-hot encoding-featurized experiments.}
\label{sifig35}
\end{figure}

\begin{figure}[H]
\centering
\includegraphics[width=1\textwidth]{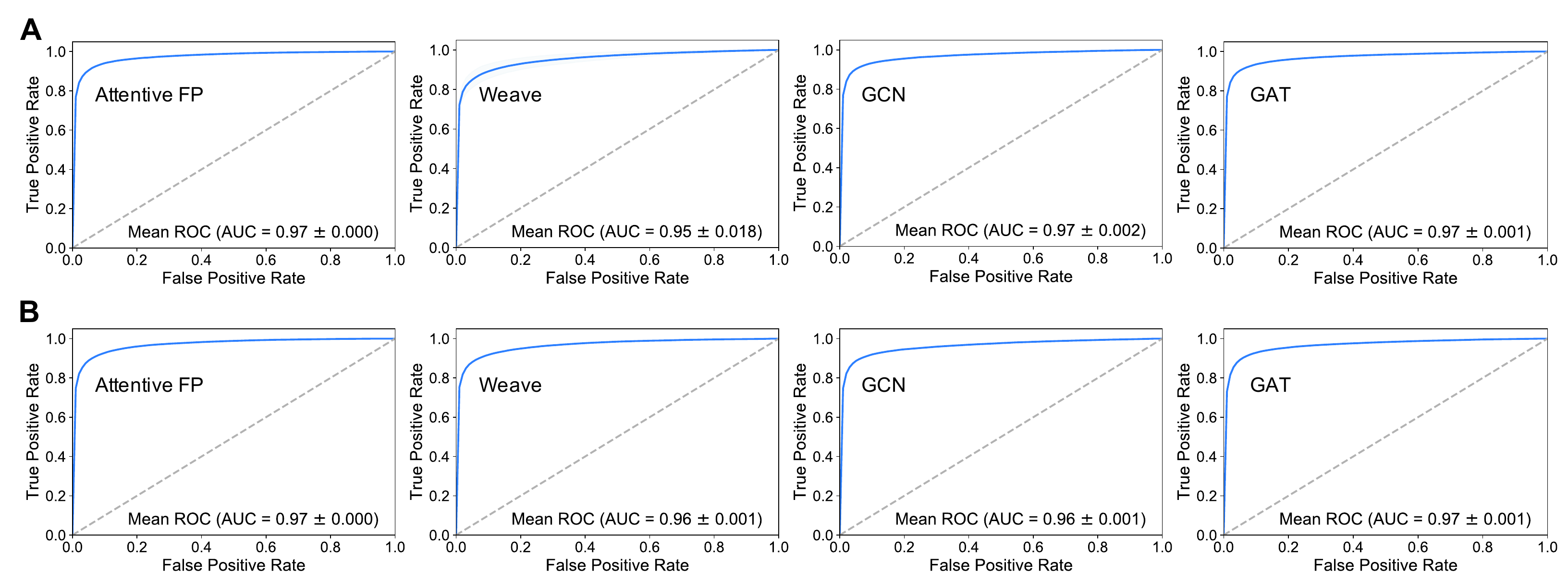}
\caption{ROC-AUC curves for classification on the species level for each of four model architectures (Weave, Attentive FP, GCN, and GAT), with the standard deviation shaded in light blue too insignificant to be visible. A standard deviation of 0.000 denotes a value of $\ <$0.001. \textbf{A.} Mean ROC-AUC curves for fingerprint-featurized experiments, with each graph displaying the mean and standard deviation across 5 sets of hyperparameters. \textbf{B.} Mean ROC-AUC curves for one-hot encoding-featurized experiments.}
\label{sifig36}
\end{figure}

\textbf{Models.} Benchmarks of multi-label classification were performed using a similar process as with immunogenicity classification for 5 different model architectures, 2 different node and edge attribution types, and 8 different taxonomic levels for a total of 80 model architecture–attribute–taxonomic level combination. For each combination, hyperparameter optimization against minimization of binary cross entropy loss was performed on SigOpt for 1000 observations and the 5 best sets of hyperparameters were extracted. Each model architecture, featurization, and taxonomic level combination was trained using the 5 best sets of hyperparameters from SigOpt using 5 distinct random seeds for splitting the dataset into train-validation-test datasets, for a total of 25 trainings per combination. The pipeline and models achieve state-of-the-art performance on multilabel taxonomic classification on all taxonomic levels (Figures \ref{sifig29}-\ref{sifig36}, Table \ref{sitab4}). Additional supplementary tables display the validation and test metrics for the most optimal set of hyperparameters, the values of the most optimal hyperparameters, and mean test metrics for each training across all 25 runs (Supplementary Tables 1-2).

\begin{table}[!ht]
\centering
\caption{The most optimal model architecture and node/edge attribute type that results in the lowest loss for prediction of all taxonomic levels. For each metric, the mean $\mu$ and standard deviation $\sigma$ are displayed across all 5 random seeds. “FP” denotes condensed fingerprint featurization, and “One-hot” denotes one-hot encoding featurization.}
\label{sitab4}
\resizebox{\textwidth}{!}{%
\begin{tabular}{@{}cccccccccccccccc@{}}
\toprule
\multirow{2}{*}{Taxonomic   Level} &
  \multirow{2}{*}{Model,   Attribute Type} &
  \multicolumn{2}{c}{ROC-AUC} &
  \multicolumn{2}{c}{F1} &
  \multicolumn{2}{c}{Recall} &
  \multicolumn{2}{c}{Precision} &
  \multicolumn{2}{c}{Accuracy} &
  \multicolumn{2}{c}{Loss} &
  \multicolumn{2}{c}{Hamming   Loss} \\ \cmidrule(l){3-16} 
        &                    & $\mu$    & $\sigma$    & $\mu$    & $\sigma$    & $\mu$    & $\sigma$    & $\mu$    & $\sigma$    & $\mu$    & $\sigma$    & $\mu$    & $\sigma$    & $\mu$    & $\sigma$    \\ \midrule
Domain  & Attentive   FP, FP & 0.993 & 0     & 0.938 & 0.002 & 0.935 & 0.004 & 0.941 & 0.001 & 0.925 & 0.001 & 0.087 & 0.002 & 0.03  & 0.001 \\ \midrule
Kingdom & Attentive   FP, FP & 0.994 & 0     & 0.912 & 0.001 & 0.891 & 0.003 & 0.934 & 0.002 & 0.892 & 0.005 & 0.056 & 0.003 & 0.019 & 0     \\ \midrule
Phylum  & GCN, FP            & 0.99  & 0     & 0.84  & 0.009 & 0.8   & 0.011 & 0.885 & 0.009 & 0.802 & 0.01  & 0.029 & 0.001 & 0.009 & 0     \\ \midrule
Class   & GCN, FP            & 0.986 & 0.001 & 0.741 & 0.008 & 0.666 & 0.01  & 0.836 & 0.006 & 0.67  & 0.009 & 0.022 & 0     & 0.007 & 0     \\ \midrule
Order   & GCN, FP            & 0.979 & 0.001 & 0.638 & 0.012 & 0.541 & 0.023 & 0.778 & 0.018 & 0.548 & 0.016 & 0.017 & 0     & 0.004 & 0     \\ \midrule
Family  & GAT, FP            & 0.975 & 0.003 & 0.557 & 0.015 & 0.456 & 0.017 & 0.715 & 0.015 & 0.469 & 0.019 & 0.013 & 0     & 0.003 & 0     \\ \midrule
Genus   & GCN, One-hot       & 0.963 & 0.003 & 0.513 & 0.015 & 0.397 & 0.015 & 0.723 & 0.020 & 0.412 & 0.010 & 0.009 & 0     & 0.002 & 0     \\ \midrule
Species & GCN, FP            & 0.968 & 0.003 & 0.487 & 0.021 & 0.382 & 0.031 & 0.675 & 0.018 & 0.395 & 0.020 & 0.007 & 0     & 0.002 & 0     \\ \bottomrule
\end{tabular}%
}
\end{table}

\subsection{Benchmarking top GNN models against results reported in literature - Glycans}
\label{si_7_4}
For the top-performing models in Tables \ref{sitab3} and \ref{sitab4}, hyperparameter optimization on SigOpt was performed again using the same method as the benchmarks in the work by \cite{Burkholz2021}: training on 80\% of the dataset and reporting metrics on the remaining 20\% used as a validation dataset (Table \ref{sitab5}). The experiment on SigOpt was optimized through minimization of the loss, and the remaining metrics (ROC-AUC, F1, recall, precision, and accuracy) were obtained as stored metrics in the SigOpt experiment. However, to obtain more robust metrics, we averaged across 25 models, instead of the 5 models in \cite{Burkholz2021}. In 4 out of 8 tasks, the models presented in our work perform better, and are comparable for the rest 4. 

\begin{table}[]
\centering
\caption{Validation metrics of the top-performing model architecture and attribute combinations for immunogenicity and taxonomic levels obtained from SigOpt using a 0.8, 0.2 train-validation split of the dataset. The subset accuracy values are compared directly with the results for augmented models presented in Table 1 of \cite{Burkholz2021}. The higher value is bolded for each task. In 4 out of 8 tasks, the models presented in our work perform better, and are comparable for the rest 4.}
\label{sitab5}
\resizebox{\textwidth}{!}{%
\begin{tabular}{@{}cccccccccc@{}}
\toprule
Class                    & Paper          & Model + Attribute Type & ROC-AUC & F1    & Recall & Precision & Accuracy       & CE Loss        & Hamming Loss \\ \midrule
\multirow{2}{*}{Immunogenicity} & This Work & Weave, one-hot & 0.999 & \textbf{0.989} & \textbf{0.989} & \textbf{0.989} & \textbf{0.989} & \textbf{0.018} & - \\ \cmidrule(l){2-10} 
                         & Burkholz, et. al. & -                      & -       & - & -  & -     & 0.942          & 0.149          & -            \\ \midrule
\multirow{2}{*}{Domain}  & This Work      & Attentive FP, fp       & 0.994   & 0.946 & 0.941  & 0.951     & 0.935 & \textbf{0.081} & 0.027        \\ \cmidrule(l){2-10} 
                         & Burkholz, et. al. & -                      & -       & -     & -      & -         & \textbf{0.939}          & 0.181          & -            \\ \midrule
\multirow{2}{*}{Kingdom} & This Work      & Attentive FP, fp       & 0.997   & 0.936 & 0.895  & 0.922     & \textbf{0.921} & \textbf{0.043} & 0.014        \\ \cmidrule(l){2-10} 
                         & Burkholz, et. al. & -                      & -       & -     & -      & -         & 0.907          & 0.301          & -            \\ \midrule
\multirow{2}{*}{Phylum}  & This Work      & GCN, fp                & 0.991   & 0.845 & 0.806  & 0.889     & 0.808 & \textbf{0.029} & 0.009        \\ \cmidrule(l){2-10} 
                         & Burkholz, et. al. & -                      & -       & -     & -      & -         & \textbf{0.841}          & 0.603          & -            \\ \midrule
\multirow{2}{*}{Class}   & This Work      & GCN, FP                & 0.987   & 0.776 & 0.713  & 0.850     & 0.718 & \textbf{0.021} & 0.006        \\ \cmidrule(l){2-10} 
                         & Burkholz, et. al. & -                      & -       & -     & -      & -         & \textbf{0.745}          & 0.976          & -            \\ \midrule
\multirow{2}{*}{Order}   & This Work      & GCN, fp                & 0.979   & 0.638 & 0.541  & 0.778     & 0.548 & \textbf{0.017} & 0.004        \\ \cmidrule(l){2-10} 
                         & Burkholz, et. al. & -                      & -       & -     & -      & -         & \textbf{0.582}          & 1.728          & -            \\ \midrule
\multirow{2}{*}{Family}  & This Work      & GAT, fp                & 0.984   & 0.617 & 0.525  & 0.748     & \textbf{0.542} & \textbf{0.011} & 0.003        \\ \cmidrule(l){2-10} 
                         & Burkholz, et. al. & -                      & -       & -     & -      & -         & 0.535          & 2.051          & -            \\ \midrule
\multirow{2}{*}{Genus}   & This Work      & GCN, fp                & 0.975   & 0.546 & 0.453  & 0.688     & 0.456 & \textbf{0.009} & 0.002        \\ \cmidrule(l){2-10} 
                         & Burkholz, et. al. & -                      & -       & -     & -      & -         & \textbf{0.475}          & 2.320          & -            \\ \midrule
\multirow{2}{*}{Species} & This Work      & GCN, fp                & 0.976   & 0.510 & 0.403  & 0.694     & \textbf{0.438} & \textbf{0.007} & 0.002        \\ \cmidrule(l){2-10} 
                         & Burkholz, et. al. & -                      & -       & -     & -      & -         & 0.428          & 2.505          & -            \\ \bottomrule
\end{tabular}%
}
\end{table}

\subsection{Peptide graphs regression}
\label{si_7_5}
\textbf{Dataset.} As explained in Appendix Section \ref{si_2_2}, the overall peptides dataset was filtered into two sub-datasets, one for E. coli and one for S. aereus. The E. coli dataset contains 4,445 peptides, including 4330 monomers, 71 multimers, and 44 multi-peptides. The S. aereus dataset contains 3,686 peptides, including 3617 monomers, 61 multimers, and 8 multi-peptides. The training for prediction of the target activity, or antimicrobial concentration, of the peptides was performed on 60\%, validated on 20\%, and tested on held-out 20\% data. 

\begin{figure}[H]
\centering
\includegraphics[width=1\textwidth]{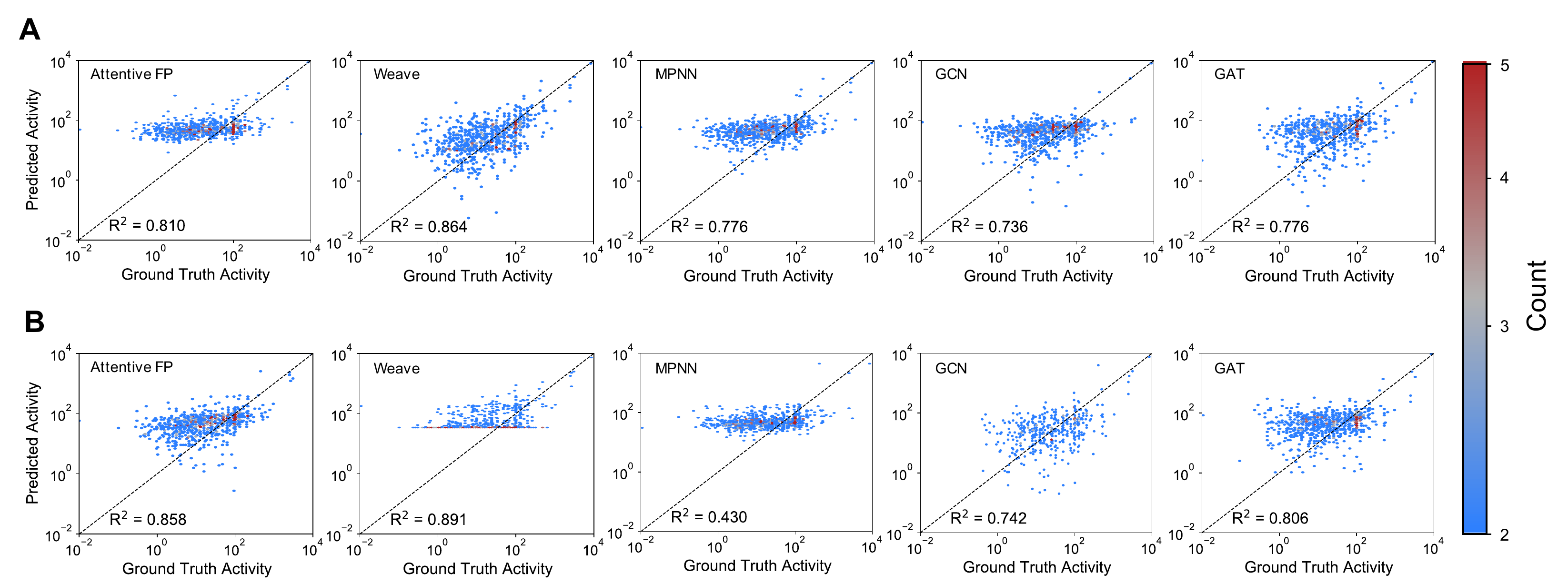}
\caption{Parity plots for \emph{Escherichia coli} validation dataset normalized with standard scaler for regression of peptide target activity for each of five model architectures (Attentive FP, Weave, MPNN, GCN, and GAT), all using the same train-validation-test dataset split. \textbf{A.} Parity plots for fingerprint-featurized experiments, with each graph displaying the top set of hyperparameters. \textbf{B.} Parity plots for one-hot encoding-featurized experiments, with each graph displaying the top set of hyperparameters.}
\label{sifig37}
\end{figure}

\begin{figure}[H]
\centering
\includegraphics[width=1\textwidth]{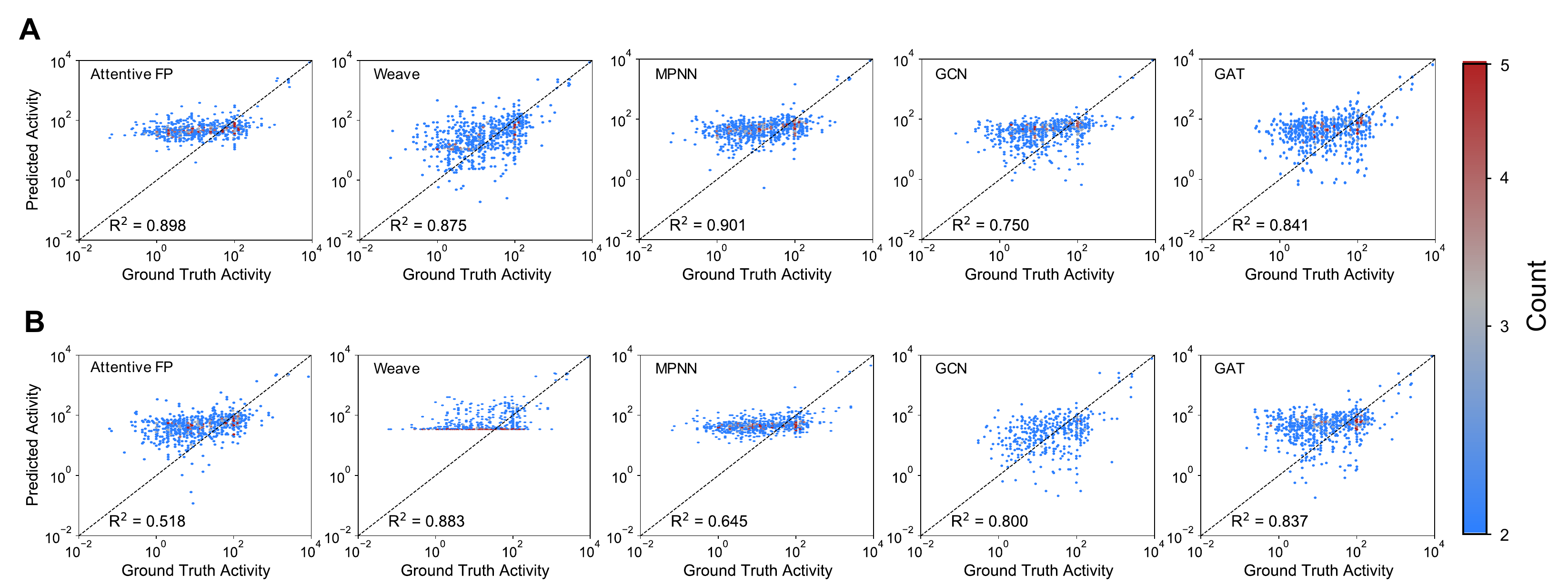}
\caption{Parity plots for \emph{Escherichia coli} test dataset normalized with standard scaler for regression of peptide target activity for each of five model architectures (Attentive FP, Weave, MPNN, GCN, and GAT), all using the same train-validation-test dataset split. \textbf{A.} Parity plots for fingerprint-featurized experiments, with each graph displaying the top set of hyperparameters. \textbf{B.} Parity plots for one-hot encoding-featurized experiments, with each graph displaying the top set of hyperparameters.}
\label{sifig38}
\end{figure}

\begin{figure}[H]
\centering
\includegraphics[width=1\textwidth]{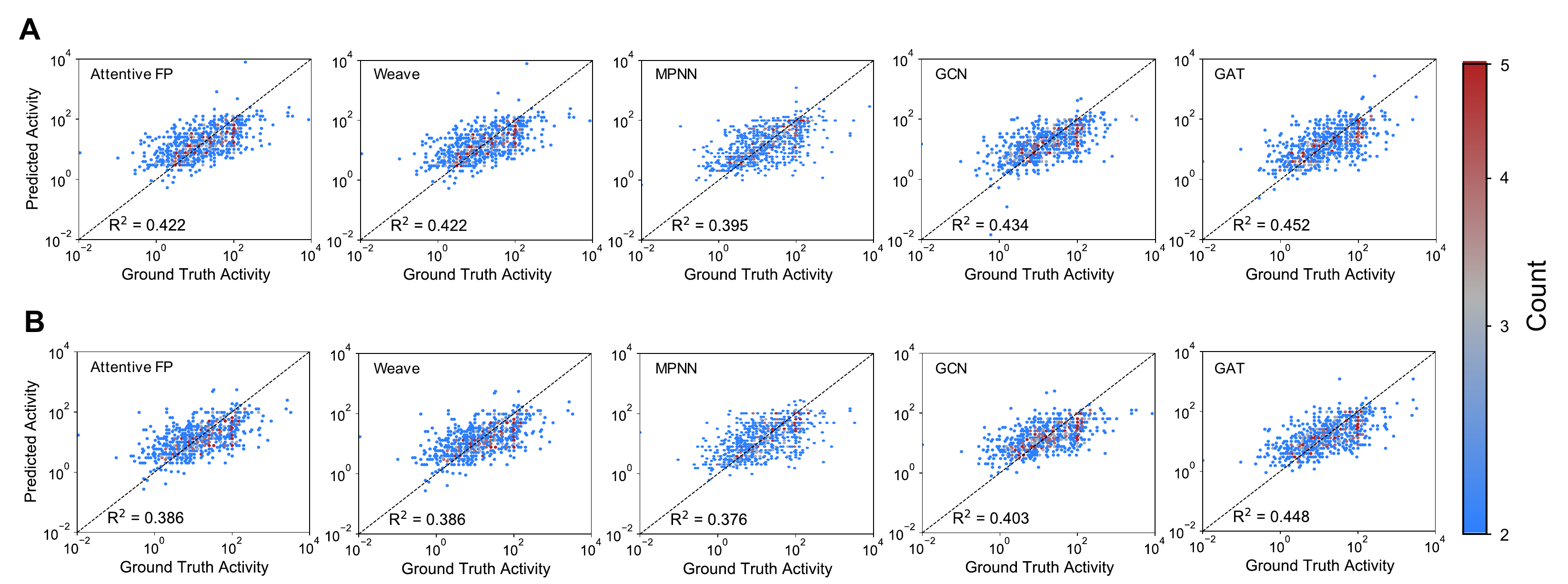}
\caption{Parity plots for \emph{Escherichia coli} validation dataset normalized with quantile transform for regression of peptide target activity for each of five model architectures (Attentive FP, Weave, MPNN, GCN, and GAT). \textbf{A.} Parity plots for fingerprint-featurized experiments, each graph for the top set of hyperparameters and the same train-validation-test dataset split. \textbf{B.} Parity plots for one-hot encoding-featurized experiments, each graph for the top set of hyperparameters and the same train-validation-test dataset split.}
\label{sifig39}
\end{figure}

\begin{figure}[H]
\centering
\includegraphics[width=1\textwidth]{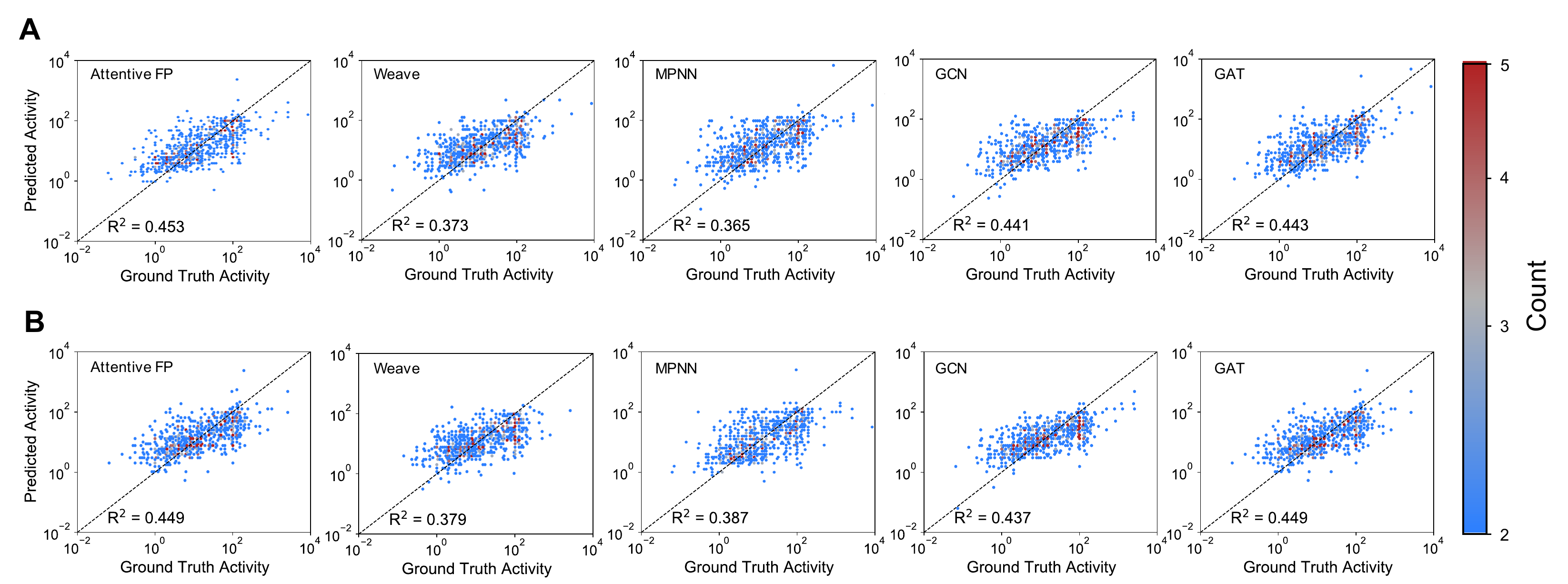}
\caption{Parity plots for \emph{Escherichia coli} test dataset normalized with quantile transform for regression of peptide target activity for each of five model architectures (Attentive FP, Weave, MPNN, GCN, and GAT). \textbf{A.} Parity plots for fingerprint-featurized experiments, each graph for the top set of hyperparameters with the same train-validation-test dataset split across all plots. \textbf{B.} Parity plots for one-hot encoding-featurized experiments, each graph for the top set of hyperparameters with the same train-validation-test dataset split across all plots.}
\label{sifig40}
\end{figure}

\begin{figure}[H]
\centering
\includegraphics[width=1\textwidth]{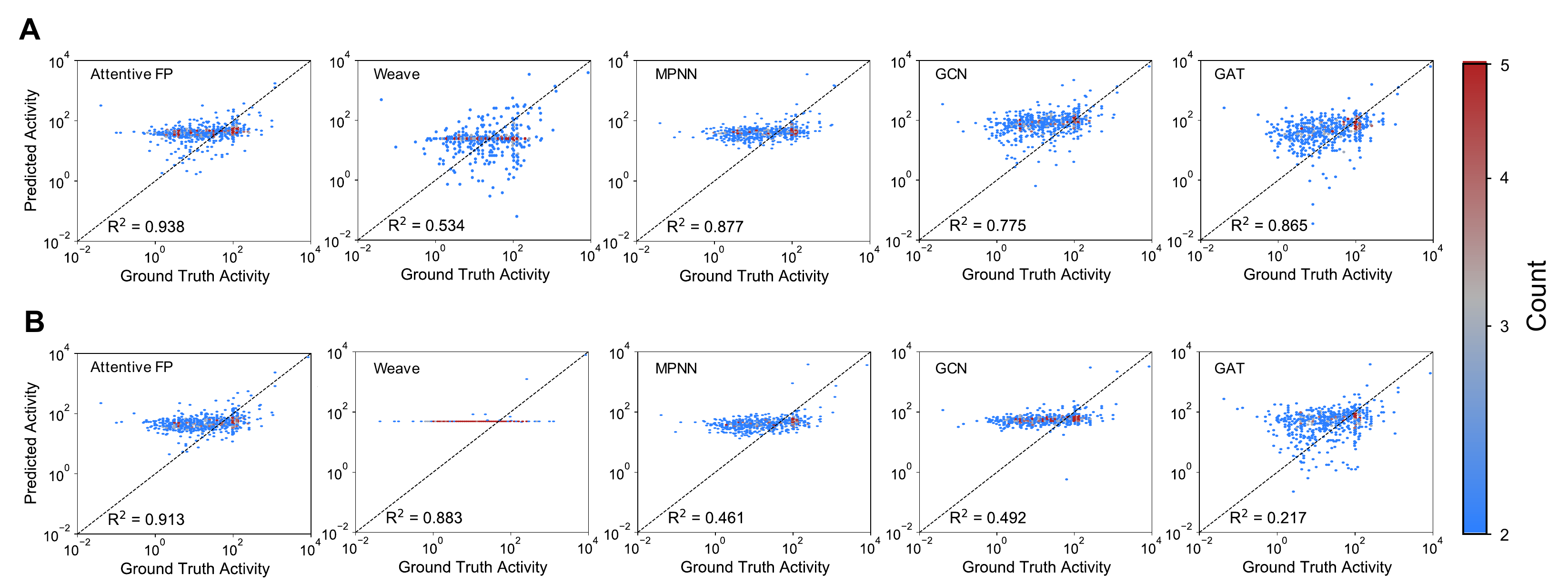}
\caption{Parity plots for \emph{Staphylococcus aereus} validation dataset normalized with standard scaler for regression of peptide target activity for each of five model architectures (Attentive FP, Weave, MPNN, GCN, and GAT). \textbf{A.} Parity plots for fingerprint-featurized experiments, each graph for the top set of hyperparameters with the same train-validation-test dataset split across all plots. \textbf{B.} Parity plots for one-hot encoding-featurized experiments, each graph for the top set of hyperparameters with the same train-validation-test dataset split across all plots.}
\label{sifig41}
\end{figure}

\begin{figure}[H]
\centering
\includegraphics[width=1\textwidth]{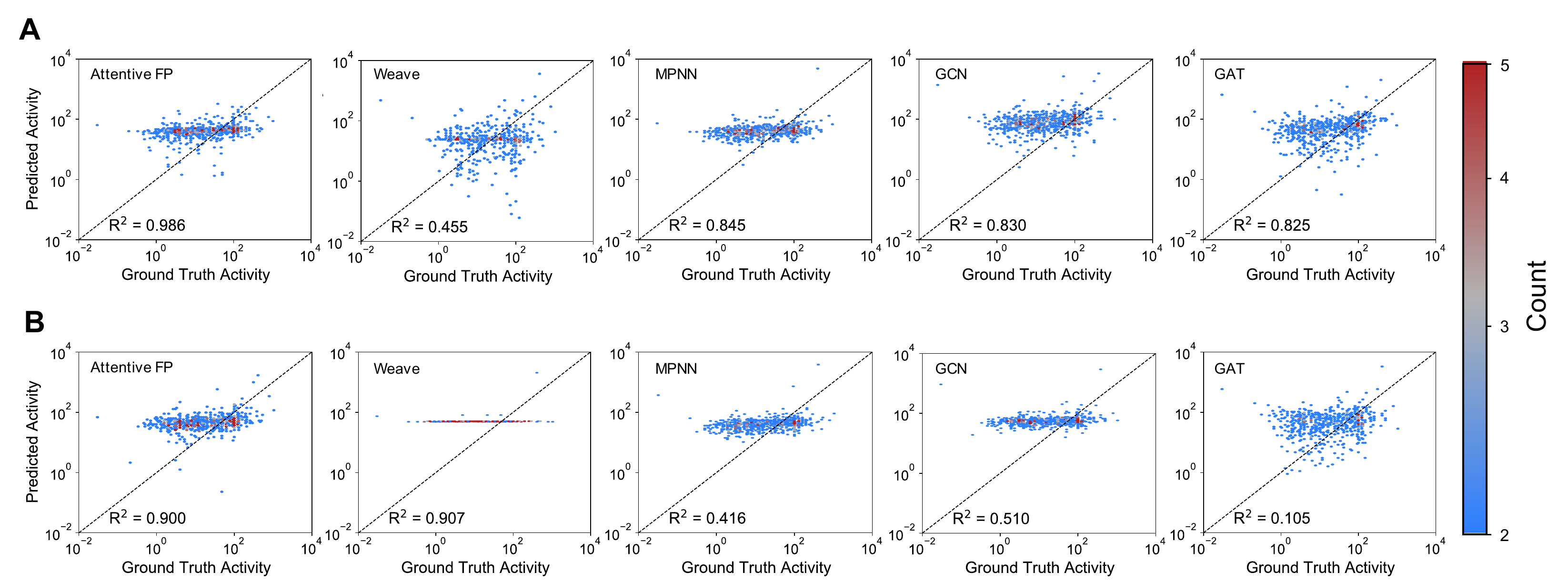}
\caption{Parity plots for \emph{Staphylococcus aereus} test dataset normalized with standard scaler for regression of peptide target activity for each of five model architectures (Attentive FP, Weave, MPNN, GCN, and GAT). \textbf{A.} Parity plots for fingerprint-featurized experiments, each graph for the top set of hyperparameters with the same train-validation-test dataset split across all plots. \textbf{B.} Parity plots for one-hot encoding-featurized experiments, with each graph displaying the top set of hyperparameters with the same train-validation-test dataset split across all plots.}
\label{sifig42}
\end{figure}

\begin{figure}[H]
\centering
\includegraphics[width=1\textwidth]{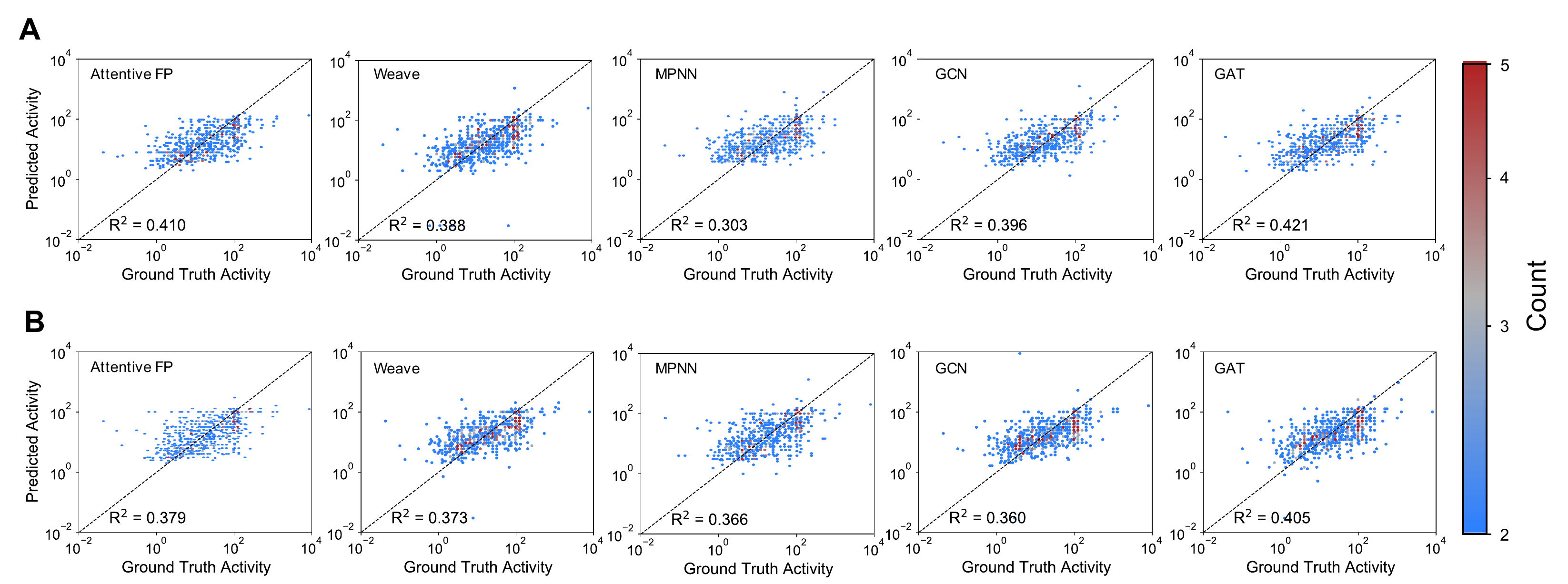}
\caption{Parity plots for \emph{Staphylococcus aereus} validation dataset normalized with quantile transform for regression of peptide target activity for each of five model architectures (Attentive FP, Weave, MPNN, GCN, and GAT). \textbf{A.} Parity plots for fingerprint-featurized experiments, each graph for the top set of hyperparameters with the same train-validation-test dataset split across all plots. \textbf{B.} Parity plots for one-hot encoding-featurized experiments, each graph for the top set of hyperparameters with the same train-validation-test dataset split across all plots.}
\label{sifig43}
\end{figure}

\begin{figure}[H]
\centering
\includegraphics[width=1\textwidth]{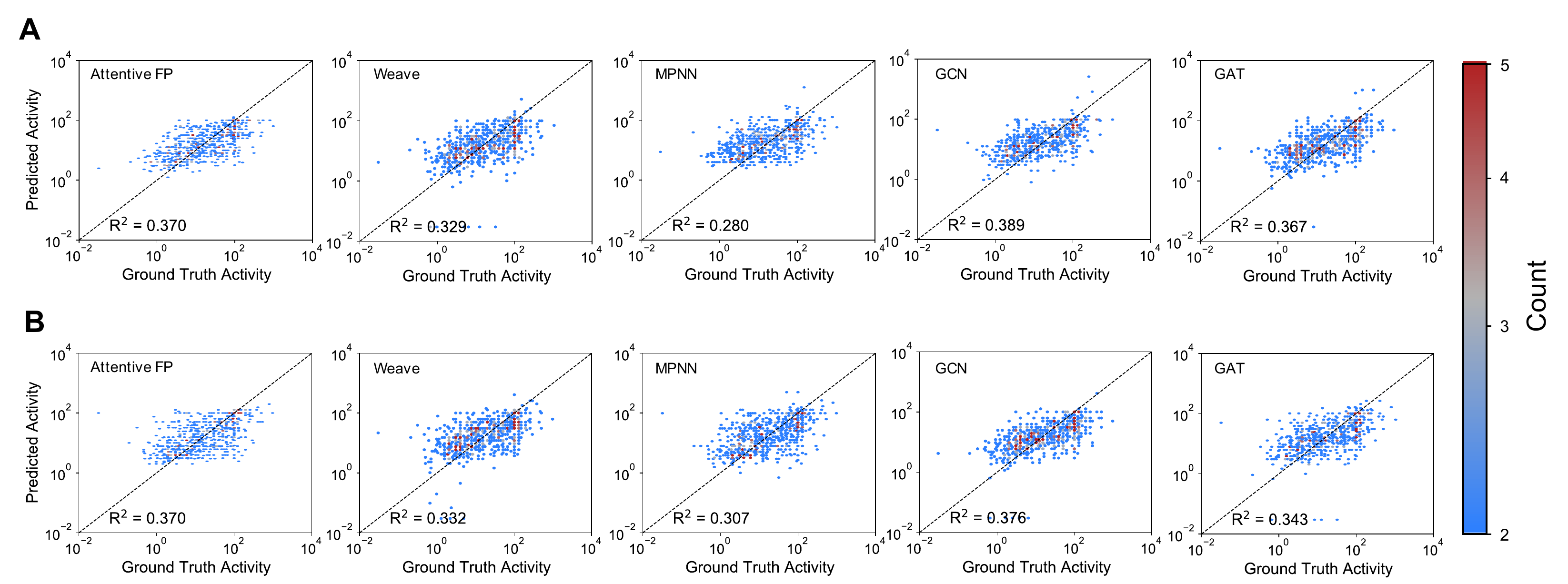}
\caption{Parity plots for \emph{Staphylococcus aereus} test dataset normalized with quantile transform for regression of peptide target activity for each of five model architectures (Attentive FP, Weave, MPNN, GCN, and GAT). \textbf{A.} Parity plots for fingerprint-featurized experiments, each graph for the top set of hyperparameters with the same train-validation-test dataset split across all plots. \textbf{B.} Parity plots for one-hot encoding-featurized experiments, each graph for the top set of hyperparameters with the same train-validation-test dataset split across all plots.}
\label{sifig44}
\end{figure}

\textbf{Models.} We performed target activity regression using 5 model architectures combined with 2 different node and edge featurization types and 2 dataset normalization types, for a total of 20 model combinations for each of the two species. The 2 normalization methods were standard scaler and quantile transform from scikit-learn \citep{Pedregosa2011}. For each benchmark, hyperparameter optimization against minimization of root mean square error (RMSE) was performed on SigOpt \citep{sigopt-web-page}, for 500 observations and the 5 best sets of hyperparameters were extracted. Each model architecture and featurization combination was trained using the 5 best sets of hyperparameters from SigOpt using 5 distinct random seeds for splitting the dataset into train-validation-test datasets, for a total of 25 trainings per model combination. All models achieve stellar performance on all metrics, with little meaningful difference in performance between fingerprint and one-hot encoding featurization (Figures \ref{sifig37}-\ref{sifig44}, Appendix Table \ref{sitab6}). Additional supplementary tables display the validation and test metrics for the most optimal set of hyperparameters, the values of the most optimal hyperparameters, and mean test metrics for each training across all 25 runs (Supplementary Tables 3-4).

\begin{table}[!h]
\centering
\caption{The most optimal model architecture, node/edge attribute type, and normalization method that results in the lowest RMSE for target activity regression. For each metric, the mean $\mu$ and standard deviation $\sigma$ are displayed across all 5 random seeds. “FP” denotes condensed fingerprint featurization, and “One-hot” denotes one-hot encoding featurization.}
\label{sitab6}
\resizebox{\textwidth}{!}{%
\begin{tabular}{@{}cccccccccccccc@{}}
\toprule
\multirow{2}{*}{Species} &
  \multirow{2}{*}{Model, Attribute, Normalization Type} &
  \multicolumn{2}{c}{Pearson r2} &
  \multicolumn{2}{c}{MAE} &
  \multicolumn{2}{c}{RMSE} &
  \multicolumn{2}{c}{Spearman r} &
  \multicolumn{2}{c}{Kendall's Tau} &
  \multicolumn{2}{c}{Loss} \\ \cmidrule(l){3-14} 
        &                    & $\mu$    & $\sigma$    & $\mu$    & $\sigma$    & $\mu$    & $\sigma$    & $\mu$    & $\sigma$    & $\mu$    & $\sigma$    & $\mu$    & $\sigma$    \\ \midrule
\emph{E. coli} & Weave, FP, Quantile transform & 0.465 & 0.023 & 0.166 & 0.004 & 0.212 & 0.006 & 0.68 & 0.019 & 0.501 & 0.014 & 0.023 & 0.002 \\ \midrule
\emph{S. aereus} & GCN, One-hot, Quantile transform & 0.38 & 0.032 & 0.185 & 0.008 & 0.229 & 0.007 & 0.62 & 0.027 & 0.451 & 0.023 & 0.026 & 0.002     \\  \bottomrule
\end{tabular}%
}
\end{table}

\section{Attribution analysis}
\label{si_8}
\subsection{Selection of model architecture - attribution method}
\label{si_8_1}
We used integrated gradients (IGs) \citep{IGaxiomatic} and Input x Grad (InpGrad) \citep{inpxgrad} for the attribution analysis of the GNNs - Weave, Attentive FP and MPNN. The model architecture selection was done to have one of each type of architecture – Weave (graph convolution), Attentive FP (graph attention), and MPNN (message passing). 

The macromolecule is represented as a graph, $\mathcal{G}(V,E)$, where V represents monomers at vertices/nodes, and E represents connecting bonds at the edges.

IGs interpolate between the input graph and a baseline graph, where all features are zero, and accumulate the gradient values for each node (Equation \ref{eqn1}). The notation follows \citep{Sanchez-lengeling2020}.
\begin{equation}
 \mathcal{G}_{A}=\left(\mathcal{G}-\mathcal{G}^{\prime}\right) \int_{\alpha=0}^{1} \frac{d y\left(\mathcal{G}^{\prime}+\alpha\left(\mathcal{G}-\mathcal{G}^{\prime}\right)\right)}{d \mathcal{G}} d \alpha
\label{eqn1}
\end{equation}
 
InpGrad is the element-wise product of the input graph and the gradient.
\begin{equation}
\mathcal{G}_{A}={\left(\frac{d \hat{y}}{d \mathcal{G}}\right)}^{T} \mathcal{G}
\label{eqn2}
\end{equation}

For the attention-based GNN, AttentiveFP, in addition to IGs and InpGrad, we evaluated attribution using attention weights, where the node attention weights are obtained by averaging over the attention scores of the adjacent nodes. 

For each attribution method, we obtained the node weights by multiplying the positive weights with the input fingerprint vectors -
\begin{equation}
\mathbf{n} = \sum_{nodes} \mathcal{G}_{A}^{+} \; \mathcal{G}, \;\;\;\;
\label{eqn3}
\end{equation}

The node weights were normalized to the maximum node weight to obtain the normalized weights.
\begin{equation}
\mathbf{n_{norm}} = \frac{\mathbf{n}}{max(\mathbf{n})}
\label{eqn4}
\end{equation}


Consistency of node weights, as defined in \citet{Sanchez-lengeling2020}, was used for evaluation of different attribution methods and model architectures.

\begin{figure}[h!]
\centering
\includegraphics[width=1\textwidth]{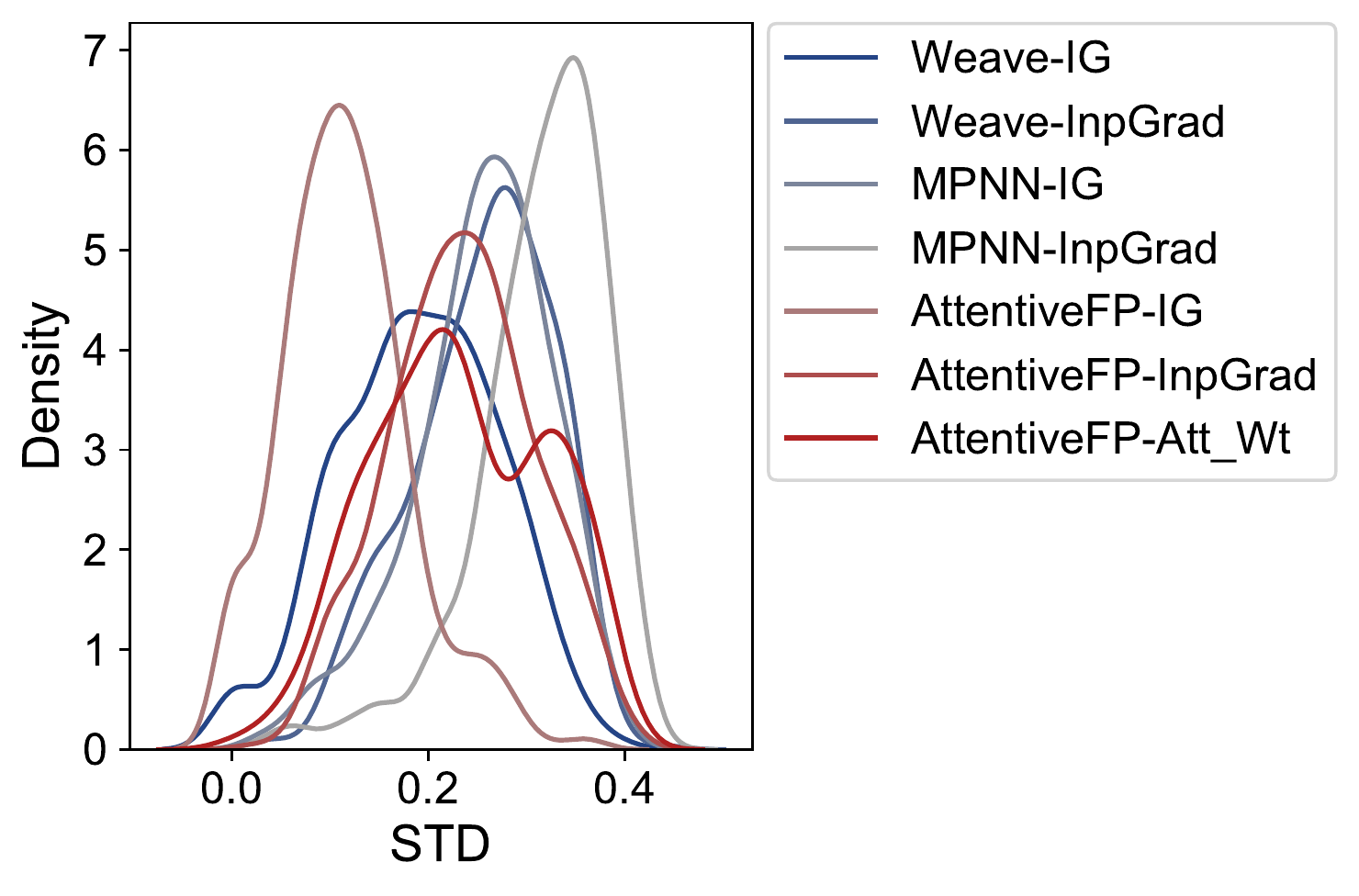}
\caption{Distribution of normalized node weights for different model architecture - attribution methods. Standard deviation of node weights obtained from 25 different model implementations – top 5 hyperparameter sets and 5 random seeds – were pooled across all immunogenic glycans to calculate the distribution for each model architecture and attribution combination.}
\label{sifig45}
\end{figure}

The distribution of standard deviation for the node weights calculated using IG and InpGrad have varying attribution consistency, both across attribution methods and model architectures (Figure \ref{sifig45}). By visual inspection, Attentive FP - IG has the smallest mean amongst all other distributions, thus the most consistent model architecture - attribution. All further attribution analysis has been done using node weights obtained from Attentive FP - IG.

\begin{table}[!h]
\centering
\caption{Top monomers contributing to glycan immunogenicity, arranged in descending order of average node weight. Monomers which occur at least 20 times in the entire data set have been included in the table. The node weights have been obtained using Attentive FP – IG combination.}
\label{sitab7}
\resizebox{0.3\textwidth}{!}{%
\begin{tabular}{@{}cc@{}}
\toprule
\multicolumn{1}{c}{\textbf{Monomer}} & \textbf{Mean Node Weight} \\ \midrule
NeuNGc                                 & 0.73                      \\ \midrule
Xyl                                    & 0.65                      \\ \midrule
Fuc                                    & 0.65                      \\ \midrule
Glc                                    & 0.64                      \\ \midrule
ManNAc                                 & 0.64                      \\ \midrule
Galf                                   & 0.63                      \\ \midrule
NeuNAc                                 & 0.62                      \\ \midrule
FucNAc                                 & 0.61                      \\ \midrule
QuiNAc                                 & 0.61                      \\ \bottomrule
\end{tabular}%
}
\end{table}

\subsection{Visualization of key substructures}
\label{si_8_2}
To visualize the responsible substructures in the monomers, we used $\mathcal{G}_{A}$ in Equation \ref{eqn1}, and multiplied the weights of the with the respective monomer fingerprint. This approach resulted in a weights vector with the same size as the fingerprint, with the most positively to the most negatively influencing substructure for the prediction. Using RDKit, we visualized the chemical substructures at different fingerprint indices and mapped it to the weights.

\subsection{Ablation analysis}
\label{si_8_3}
\begin{figure}[H]
\centering
\includegraphics[width=1\textwidth]{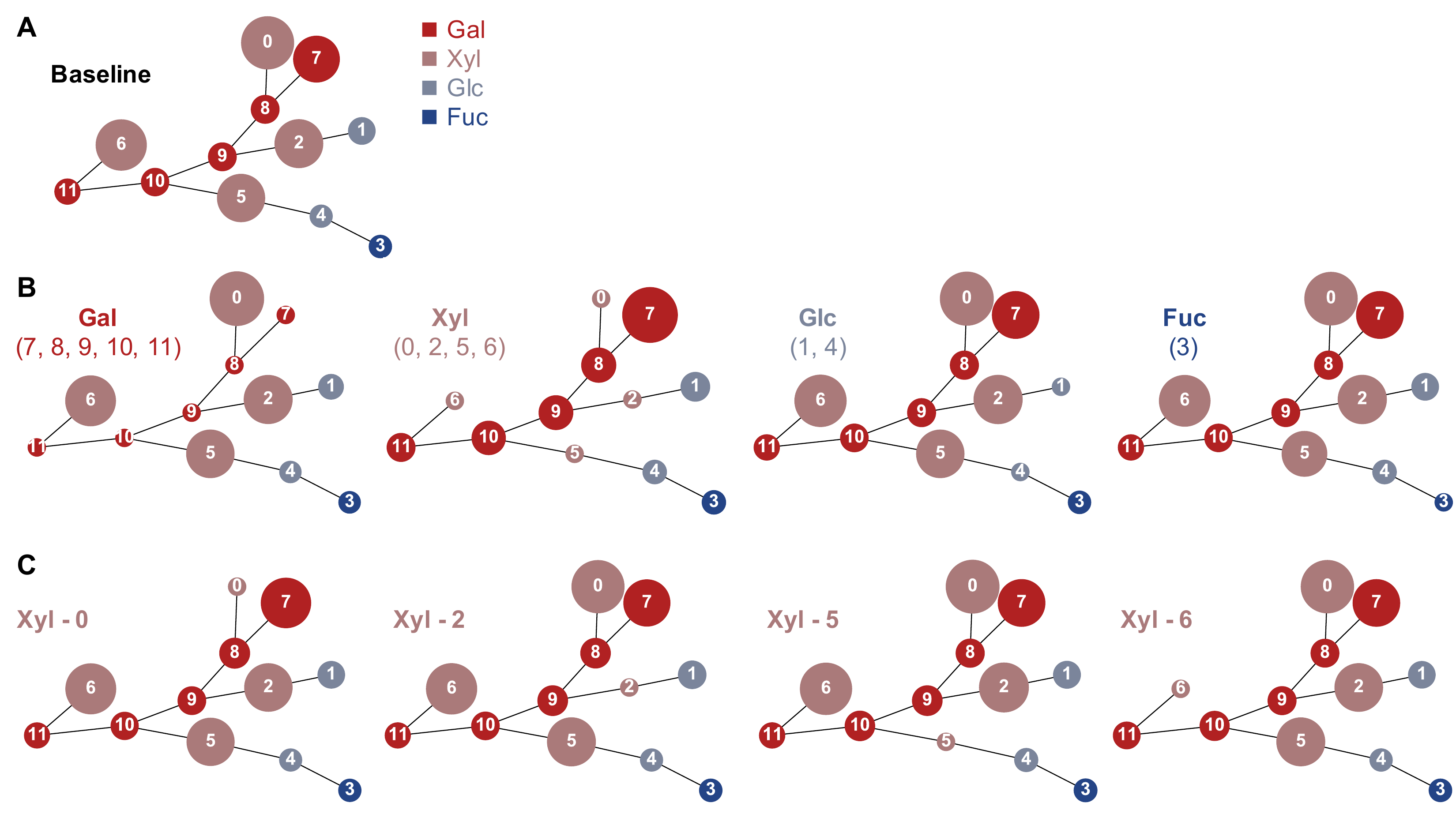}
\caption{Ablation analysis for immunogenic glycans shows that the importance of non-ablated monomers remains consistent, irrespective of node- and monomer-specific ablation. \textbf{A.} Glycan without any ablations is visualized as the baseline case. \textbf{B.} Monomer-specific ablation shows that ablation of a particular monomer has a negligible impact on the importance/weight of other monomers. Ablation was done by assigning zero tensors to nodes of respective monomers. This experiment was done by subsequently mutating all nodes corresponding to Gal, Xyl, Glc, and Fuc. \textbf{C.} Node-specific ablation shows that other nodes of a particular monomer have a negligible change when one of the nodes is assigned a zero tensor. This experiment was done by mutating all 4 nodes corresponding to Xyl, one at a time.}
\label{sifig46}
\end{figure}

We performed ablation analysis to check if the monomer importance or node weights changed when we assigned a zero tensor to one or more nodes (Figure \ref{sifig46}). For node- and monomer type-specific ablation, we observed that the importance of non-ablated nodes remained consistent. 

\vfill
\pagebreak
\section*{Supplementary Tables}
\href{https://www.dropbox.com/s/1cybiwarhppmqcg/SupplTable1.xlsx?dl=0}{Supplementary Table 1}: Validation and test metrics for the top performing sets of model hyperparameters averaged across model implementations with 5 random seeds for glycan classification, as well as test metrics averaged across 25 model implementations (5 hyperparameter sets and 5 random seeds)

\href{https://www.dropbox.com/s/qz3l53siu6bdw61/SupplTable2.xlsx?dl=0}{Supplementary Table 2}: GNN model hyperparameter sets for the top performing glycan classification models 

\href{https://www.dropbox.com/s/0hqaixghu6ps6zk/SupplTable3.xlsx?dl=0}{Supplementary Table 3}: Validation and test metrics for the top performing sets of model hyperparameters averaged across model implementations with 5 random seeds for peptide regression, as well as test metrics averaged across 25 model implementations (5 hyperparameter sets and 5 random seeds)

\href{https://www.dropbox.com/s/oavnd8a2wkh7xdd/SupplTable4.xlsx?dl=0}{Supplementary Table 4}: GNN model hyperparameter sets for the top performing peptide regression models

\end{document}

%% file: math_commands.tex
\usepackage{amsmath,amsfonts,bm}

\def\eqref#1{equation~\ref{#1}}

\def\1{\bm{1}}

\DeclareMathAlphabet{\mathsfit}{\encodingdefault}{\sfdefault}{m}{sl}
\SetMathAlphabet{\mathsfit}{bold}{\encodingdefault}{\sfdefault}{bx}{n}

